# Airport take-off and landing optimization through genetic algorithms

Fernando Guedan-Pecker[1] | Cristian Ramirez-Atencia[2]

[1]Iptima Herencias, Madrid, Spain

[2]Department of Computer Engineering Systems, Universidad Politecnica de Madrid, Madrid, Spain

**Correspondence**
Cristian Ramirez-Atencia, Department of Computer Engineering Systems, Universidad Politecnica de Madrid, Madrid, Spain.
Email: cristian.ramirez@upm.es

Funding Information
This research was supported by the Spanish Ministry of Science and Education, FightDIS project (PID2020-H7263GB-100, 2021-2023); and European Commission, IBERIFIER project (2021-2023): Iberian Digital Media Research and FactChecking Hub (CEF-TC-2020-2: 2020-EU-IA-0252).

**Abstract**

This research addresses the crucial issue of pollution from aircraft operations, focusing on optimizing both gate allocation and runway scheduling simultaneously, a novel approach not previously explored. The study presents an innovative genetic algorithm-based method for minimizing pollution from fuel combustion during aircraft take-off and landing at airports. This algorithm uniquely integrates the optimization of both landing gates and take-off/landing runways, considering the correlation between engine operation time and pollutant levels. The approach employs advanced constraint handling techniques to manage the intricate time and resource limitations inherent in airport operations. Additionally, the study conducts a thorough sensitivity analysis of the model, with a particular emphasis on the mutation factor and the type of penalty function, to fine-tune the optimization process. This dual-focus optimization strategy represents a significant advancement in reducing environmental impact in the aviation sector, establishing a new standard for comprehensive and efficient airport operation management. The results obtained after the fine-tuning showed that our algorithm is able to obtain a solution with pollution emissions only 0.55% higher than the optimal solution within a limited execution time.

## 1 | INTRODUCTION

The main air pollutants produced by aircraft engines from fuel combustion are nitrogen oxides (NOx), hydrocarbons (HC), and fine particulate matter (PM). Although airplanes traveling at different altitudes generate emissions that have an impact on air, the main impacts occur around airports and are associated with landing and take-off (LTO) operations.

A flight can be divided, according to fuel consumption and consequently pollution, into 3 phases (Winther & Rypdal, 2020):

- Departure: Activities near the airport carried out below 3,000 ft altitude. It includes taxi-out, take-off and climb-out operations.
- Cruise climb and descent (CCD): Activities above 3,000 ft, including climb operations, cruise phase and the descent from cruise altitude to 3,000 ft.
- Arrival: Phases near the airport and below 3,000 ft, including final approach, landing and taxi-in.

LTO operations include Departure and Arrival (see Figure 1), with average operation times, according to ICAO, of 2.9 min and 4.0 min, respectively (see Table 1).

The International Civil Aviation Organization (ICAO) has established in 32.9 minutes the average operating times for LTO operations (Winther & Rypdal, 2020, pag 12).

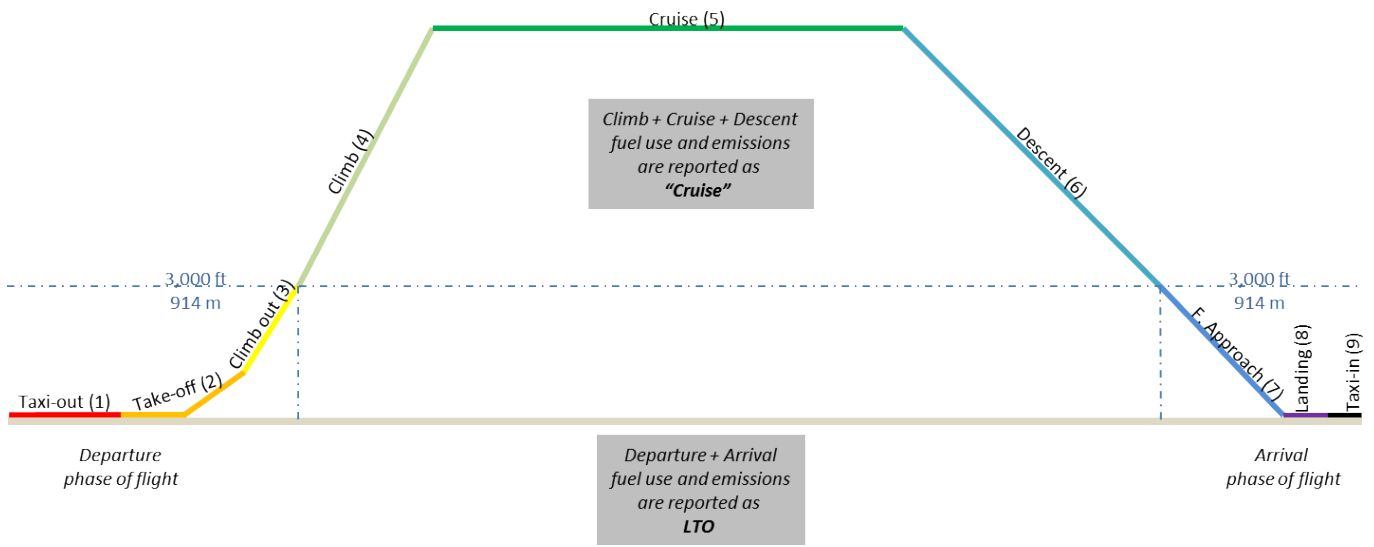

**FIGURE 1** Flight Phases from EMEP/EEA air pollutant emission inventory guidebook

**TABLE 1** Average operating times for LTO operations according to ICAO

| Take-off | Climb-out | Approach-LAN | Taxi/ground |
| --- | --- | --- | --- |
| 0,7 min | 2,2 min | 4,0 min | 26,0 min |

According to data recorded by 'Airport Tracker'[‡], web portal created by the Overseas Development Institute (ODI), the European Federation of Transport and Environment (T&E) and the International Council on Clean Transport (ICCT), Dubai Airport leads the ranking of $CO_2$ emissions (16.6 million tons), followed by London Heathrow (16.2), Los Angeles (15.3), John F. Kennedy in New York (15.3) and Paris-Charles de Gaulle (11.5).

Taking into account that the population's awareness of its harmful effects has evolved very significantly in recent years, this work evaluates the optimization of aircraft LTO activities at an airport to minimize the polluting effects produced by fuel combustion.

To obtain the optimal configuration described above, Genetic Algorithms (GA) have been used as they allow to solve efficiently complex problems where the duple 'capacities/requirements' are tight. A special study has been carried out on the gen codification as well as the fitness function, which control the levels of airplane contamination. As the problem resolution requires boundary conditions, we have included in the GA model constraint handling techniques.

The main contributions of this work are as follows:

1. Resolution of the LTO optimization problem by scheduling both boarding gates and runways at the same time for arrivals and departures. Up to our knowledge, there is no other contribution in the state-of-the-art that considers this dual focus.
2. Use of a Genetic Algorithm (GA) method to minimize the air pollution (measured in terms of polluting time) in the LTO optimization problem.
3. Sensitivity study of different mutation criteria and constraint handling techniques for the GA to solve the LTO optimization problem.

This paper is structured as follows: Section 2 provides the state-of-the-art on how the problem is solved nowadays. In section 3 we adapt the GA for the LTO optimization. In section 4 the results of the model are applied to a specific configuration (New York's JFK airport ATMs corresponding to August 5, 2019). A Sensitivity Study has been added to evaluate the differences and similarities between models with different hyperparameters. Finally, section 5 shows some conclusions and future model improvements.

---

[‡] airporttracker.org

## 2 | STATE OF THE ART

From the analysis of current papers published in the aeronautical world, the vast majority of them deal with the problem of finding a time allocation vector in such a way that it satisfies certain criteria that minimize the possible joint negative impacts on the problem stakeholders. In general terms, it is known as "Strategic Airport Slot Allocation" (Katsigiannis, Zografos, & Fairbrother, 2020).

The approach to the problem is carried out for purely economic purposes, that is, to solve with current means the peaks in air traffic that are expected to occur in the coming years. According to Jacquillat and Odoni (Jacquillat & Odoni, 2018) capacity/demand problems are usually analyzed from a triple point of view: (i) increase in capacity through the construction of new airports or the increase/improvement of existing ones, (ii) improvement of operational techniques and (iii) actions on demand management (control of overcapacity generated at congested airports during peak days).

The most widespread way to control this congestion problem is to apply differential rates based on the assigned slot time. Currently and outside of the USA, IATA (International Air Transport Association) is in charge of solving the assignment of slots between airports with high levels of congestion, given that the slots of one airport affect those of another: the departure slot of an aircraft from the origin airport, added to the flight time, implies establishing an arrival slot at the destination airport in the required window time (Androutsopoulos & Madas, 2019).

From the point of view of the mathematical methodology used to solve allocation problems, almost all studies used linear programming based on the branch-and-bound method(Lohatepanont & Barnhart, 2004; K. Wei, Vaze, & Jacquillat, 2020), and some works have extended the approach using rolling horizon model(Abdelghany & Abdelghany, 2008), Lagrangian Relaxation(Gomez Miguel, 2019) or Large Neighborhood Search(Androutsopoulois, Manousakis, & Madas, 2020), among other methods. This work scheme presents lights and shadows. The lights are the ability to include a large number of constraint equations that, when solving the model, are guaranteed their perfect compliance. The shadows are the large number of binary variables that need to be implemented when generating the model for resolution. It is this condition that makes models with large numbers of variables practically unsolvable without high computing capacities. There are two main programs used for solving these linear programming approaches: IBM ILOG CPLEX Optimization Studio (K. Wei et al., 2020; Lohatepanont & Barnhart, 2004; Androutsopoulois et al., 2020; Ribeiro, Jacquillat, Antunes, Odoni, & Pita, 2018) and Gurobi Optimizer (Gomez Miguel, 2019; Katsigiannis et al., 2020; Katsigiannis & Zografos, 2021). Apart from this work scheme, there are very few works that use other approaches, such as GAs(Abdelghany & Abdelghany, 2017) to deal with the slot allocation problem.

Another different approach that has also gained some importance in recent years is the "Airport Ground Movement Problem", consisting on the routing and scheduling of different ground movement operations of aircrafts. This problem commonly focuses on one aircraft, instead of all the aircrafts working at the same time in an airport. In Adacher et al. (Adacher, Flamini, & Romano, 2018), the problem is solved using an alternative graph model and a local greedy heuristic algorithm to schedule taxiway cross points, runway intersections, parking stands, and stop bars minimizing both the total routing taxiing delay and the pollution emission in terms of total waiting time the engines are turned on. In Li et al. (N. Li et al., 2019), they use a genetic algorithm, where the chromosome represents paths, to solve the problem the same problem but considering steering too. In Nogueira et al. (Nogueira, Aguiar, Weigang, et al., 2014) they use and ant colony optimization (ACO) approach. In Murrieta et al. (Murrieta-Mendoza & Botez, 2020), several metaheuristics were compared, but they only focus on solving one flight, and then the next flight based on the already planned before, instead of solving them simultaneously. In Zhang et al. (Zhang, Huang, Liu, & Li, 2019), a multi-objective approach called NSGA-II is used to minimize taxiing time and fuel consumption. Tian et al. (Tian, Wan, Han, & Ye, 2018) also used NSGA-II to minimize pollution, noise and delays, but focusing only on arrivals, scheduling descent route, holding time and STAR route (CDA). In a posterior work by Tian et al. (Tian, Wan, Ye, Yin, & Xing, 2019), they added an Analytic Hierarchy Process (AHP) for the priority ordering of flights.

On the other hand, there are recent studies focused on the gate allocation problem(Daş, Gzara, & Stützle, 2020), in which flights are assigned to airport gates. These works are focused on different objectives to optimize, which can be passenger-oriented (minimizing walking distances, waiting or transfer times, baggage distance, etc.), airport/airline-oriented (minimizing number of flights assigned to remote gates, towing cost, waiting delays, etc.), or robustness-oriented (minimizing range of idle times, expected number of gate conflicts, etc.).

Although most studies use integer lineal optimization methods such as branch and bound(Maharjan & Matis, 2012; Neuman & Atkin, 2013; Yu, Zhang, & Lau, 2016), some recent works use metaheuristic approaches such as Simulated Annealing(Zhao & Duan, 2021), GAs(Hu & Di Paolo, 2008; Ghazouani, Hammami, & Korbaa, 2015; Chow, Ng, & Keung, 2022), Particle Swarm Optimization (PSO)(Deng et al., 2017), ACO(Deng, Sun, Zhao, Li, & Wang, 2018), Bee Colony Optimization (BCO)(Marinelli, Dell'Orco, & Sassanelli, 2015), Multi-Objective Evolutionary Algorithms(Mokhtarimousavi, Talebi, & Asgari, 2018) or even combinations of GA and fuzzy models(D.-x. Wei & Liu, 2009), and GA and bayesian models(Bagamanova & Mota, 2020) to solve the airport gate allocation problem.

Despite being less common, due to most airport having just one or two runways, there are also some studies that focus on runway scheduling, using Mixed Integer Linear Problem (MILP) modeling techniques(Clare & Richards, 2011), but also solving this problem with other metaheuristics such as Tabu Search(Soykan & Rabadi, 2016) minimizing tardiness.

However, it is surprising how little attention is paid in these articles to LTO operations and their polluting effect(Maharjan & Matis, 2012) (only a few of the Airport Ground Movement studies focused on air pollution). The relevance of the present work is the focus on the assignment of airport gates and runways at the same time for each LTO operation with the objective of minimizing the polluting effects produced by these operations.

# 3 | GENETIC ALGORITHM FOR LTO OPTIMIZATION

The objective of this work is to develop an application that, through the use of Genetic Algorithms (GA), allows establishing the assignment of airport operations, the so-called Air Traffic Movements (ATM), to the corresponding boarding gates (BG) and to the landing (LAN) and take-off (TOF) runways (RNW) in such a way as to minimize the polluting effect of said operations. It is not within the scope of the work to schedule arrival/departure times. It will be understood that the said vector of times is provided by the corresponding airport authority, which establishes the slot cadence and assigned airport terminal to the different daily airport operations.

To solve the problem, a GA model has been developed that responds to the challenge of efficiently forecasting the assignment of BGs/RNWs including boundary conditions. This last point is one of the main objectives of this work, how to include the boundary conditions through the constraint equations. In the following subsections, the airport LTO optimization problem is explained, as well as how it has been modeled with a GA.

## 3.1 | Airport LTO Optimization Problem

The objective of the LTO optimization problem is to obtain a scheduling of BG and RNW for the different ATMs of an airport in such a way that contamination levels are minimal. Within the complex general configuration of an airport, the parameters strictly necessary for the resolution of this problem are the following:

- The number of ATM operations
- The number of RNW (maximum of 9 RNW per airport)
- The number of TERminals (maximum of 9 TER per airport)
- The number of BG for each terminal (maximum of 99 BG per terminal)
- The distances $d(BG, RNW)$ between each BG and RNW track header.

Furthermore, it is necessary to measure the pollution levels generated by each of the scheduled ATMs over the considered period of time, which in this model is set to 24 hours. It is clear that the polluting factor is the combustion of fuel by aircraft engines. Comparatively, among the different types of aircraft, the higher the efficiency of the engines, the less pollution in the same unit of time. In order to facilitate the analysis of the data, the relative level of contamination of the different airplanes has been considered as a working parameter. For this, a contamination base level has been established, assigning higher/lower relative values to airplanes with high/low levels of contamination (see Table 5). These levels of contamination are calculated on the basis of the rate of different pollutant emissions produced by the airplane.

Although the levels of pollutants emitted by aircraft engines during their LTO operations could be a good measure to minimize, it has not been considered the most optimal, as it makes it difficult to understand the values obtained. Therefore, it has been decided to use the times used by airplanes in their airport operations as a unit of measurement of pollution. After all, there is a linear relationship between the pollutants emitted and the time and level of engine operation.

This unit of measurement (minutes-pollution) represents a very simple interpretation of the results obtained. The shorter the total operational time, the less pollution is emitted. Likewise, the selected unit of measurement allows inclusion in the calculation of both the operational times and the levels of individual contamination of the different aircrafts used.

## 3.2 | GA Representation

GAs are a type of optimization algorithm inspired by the process of natural selection(Katoch, Chauhan, & Kumar, 2021). They are commonly used in the field of artificial intelligence to solve complex problems that are difficult or impossible to solve using traditional methods. For the LTO optimization problem, the algorithmic scheme of the GA that has been developed to solve the problem is presented in Figure 2. The algorithm works by starting with a population of candidate solutions(Cahmbers, 2001). The fitness of each solution is evaluated (in this case, checking both the objective and constraint equations of the problem, as explained in section 3.3), and the most fit individuals are selected to produce the next generation of solutions. Then, using the principles of reproduction (crossover) and mutation, new solutions are generated. These operators are detailed in section 3.4. Finally, the new population is built based on some replacement method, which can directly replace parents with children or combine them and select the best individuals. The process is repeated iteratively until a stopping criterion is met. Over time, this process converges to an optimal solution that best satisfies the constraints and objectives of the problem.

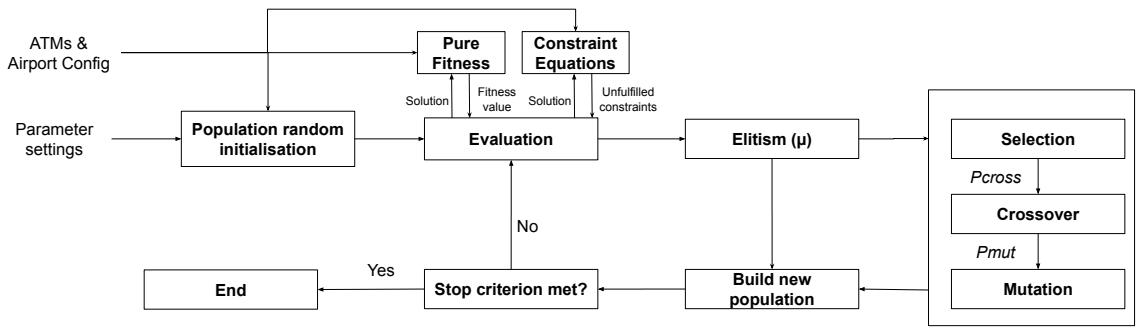

**FIGURE 2**  Scheme of the GA applied to solve the LTO optimization

One of the main aspects for working with a GA is the representation of a solution (or individual of the population). Each individual is represented by a chromosome with a number of genes, where each gene is made up of alleles. In the LTO optimization problem, the chromosomes will be made up of a number of genes equal to the number of ATM operations scheduled.

In the existing literature on GA, binary coding is recommended, which exclusively allows the use of 0 and 1 to represent the different values. In the present case, said encoding was, from the beginning, discarded due to the length that the gene would need to be to be able to contain all the required information.

After different evaluations, it was found that the encoding that best suited the objective pursued, both for simplicity and for compression, was made up of 5-digit integer genes with the following structure:

- first digit - RNW number assigned to LAN operation (0 if none).
- second digit - RNW number assigned to the TOF operation (0 if none).
- third digit – number of TERminal.
- fourth and fifth digits - BG number of the assigned Terminal.

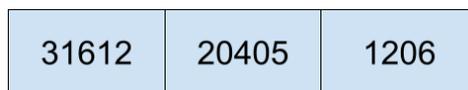

**FIGURE 3**  Example of chromosome with 3 ATMs scheduled in LTO optimization

For better understanding, in Figure 3, an example of a chromosome with 3 ATMs is presented. In this example, the 3 genes correspond to the following assignments:

31612   The LAN operation will be carried out through RNW 3 and the TOF through RNW 1. The ATM is assigned to BG 12 of Terminal 6.
20405   The LAN operation will be carried out through RNW 2 without TOF operation. The ATM is assigned to BG 5 of Terminal 4 (the assigned BG will be occupied from the LAN moment until the end of the period).
 1206   The TOF operation will be carried out through RNW 1 without LAN operation. The ATM is assigned to BG 6 of Terminal 2 (the aforementioned BG is occupied since the beginning of the period until the TOF operation).

When generating the initial population and in order to guarantee the integrity of the chromosomes, the assignment of the values of the different genes must fulfill the intervals of variation of each of the alleles in such a way that the subsequent reconstruction procedures will not proceed. For this purpose, each allele is randomly selected within its corresponding valid range of values.

## 3.3 | Fitness function

The fitness function, also called the objective function, is the function that GA seeks to optimize through its evolutionary process. Through the values provided by the function, the GA is capable of structuring the selective and evolutionary process. For each individual, fitness is calculated by adding the particular fitness value of each of the genes that form its chromosome. Therefore, the objective of the GA in LTO optimization is to minimize the value of the fitness function since said value supposes the level of contamination, expressed in minutes-contamination, of the ATMs included in a given chromosome.

In this particular GA model, it is not enough to obtain only the minimum of the aforementioned fitness function, but it is also compulsory that the structure of the chromosome complies with a series of boundary conditions, imposed through the constraint equations, that give coherence to the obtained solution. It is for this reason that a difference must be made between the concept of Total Fitness and Pure Fitness.

### 3.3.1 | Pure Fitness

The equation that measures the level of contamination (Pure Fitness) generated by each of the ATM operations that make up a given individual is given by equation 1.

$$F_p(Ind) = \sum_{i=1}^{\#ATM} \left[ \frac{d(RL_i, BG_i) + d(RT_i, BG_i)}{S_{Taxi}} \cdot 0.06 + T_L(RL_i) + PB(RT_i) + T_T(RT_i) \right] \cdot PF_i \qquad (1)$$

where *Ind* represents the individual of the population being evaluated and *#ATM* represents the number of ATM operations. $RL_i$ is the runway assigned to the LAN operation, $RT_i$ is the runway assigned to the TOF operation, and $BG_i$ is the BG assigned to ATM operation $i$. The function $d$ returns the distance, in meters, between the head of the runway and the BG. $S_{Taxi}$ is the speed, in Km/h, of the airplanes in Taxiway operations, which is assumed constant for any type of aircraft, and the number $0.06$ in the equation is used to convert from Km/h to m/min. $T_L$ is the time, in minutes, that corresponds to the approach and landing operations on the specified runway, while $T_T$ corresponds to the time for take-off and climb-out operations. *PB* returns the time, in minutes, used in the aricraft pushback operation if the TOF operation is performed. Finally, $PF_i$ is the pollution factor of the aircraft used in ATM operation $i$ according to its typology (see Table 5).

### 3.3.2 | Constraint Equations

In addition to the Pure Fitness function, and in order to be able to assign BG and RNW consistently, it is necessary to impose a series of boundary conditions on the GA model which are implemented through their respective constraint equations.

Regarding the assignment of ATMs to the different BGs, the following constraints are considered:

**CE-BG01** controls, for every BG, that the LAN/TOF sequence is logical, i.e. that a LAN operation is not assigned to a specific BG when this BG is already occupied by the previous ATM. The result of the constraint equation will be given by the number of LAN/TOF sequences out of order, i.e. the number of pairs ($k,i$) that do not satisfy equation 2.

$$\forall k \in ATM \nexists i \in ATM \quad i \neq k \land BG_i = BG_k \land (SL_k < SL_i < ST_k \lor SL_k < ST_i < ST_k) \qquad (2)$$

where $[SL_i, ST_i]$ are, respectively, and according to the scheduled time, the LAN and TOF sequence number of the ATM operation $i$.

**CE-BG02** guarantee for each of the ATMs with a single operation, either LAN or TOF, that the sequence of assignments is respected.

$$\forall k \in ATM \quad SL_k = 0 \implies \nexists i \in ATM \quad i \neq k \land BG_i = BG_k \land SL_i < ST_k \qquad (3)$$

$$\forall k \in ATM \quad ST_k = 0 \implies \nexists i \in ATM \quad i \neq k \land BG_i = BG_k \land SL_k < SL_i \qquad (4)$$

**CE-BG03** guarantees that none of the BGs is assigned a number of ATMs greater than an established limit. If this constraint is not imposed, the model, guided by the search for minimum distances to the runway headers, which is the main contamination factor, would assign a very high number of ATMs at the closer BGs. The result of the constraint will be given by the number of ATMs assigned to a specific BG that exceeded the established limit *MaxBG* in equation 5.

$$\forall b \in BG \quad \#\{i \in ATM | BG_i = b\} \leq MaxBG \qquad (5)$$

On the other hand, regarding the assignment of ATMs to the RNW for LAN and TOF, the following constraints are considered:

**CE-RNW01** guarantees, for the correct assignment of RNW, that certain aircraft, due to their size, can only use RNWs with length enough to guarantee LAN/TOF (see the typology column in Table 5). This constraint will be imposed when generating the initial population, as well as in the mutation and crossover processes, through equation 6.

$$\forall k \in ATM \quad RL_k \in AR_k \land RT_k \in AR_k \tag{6}$$

where $AR_k$ is the set of runways allowed for ATM $k$ based on the typology of the aircraft used.

**CE-RNW02** controls that there will not be, above a certain value, continuous assignments of ATMs, either LAN or TOF, to a certain RNW. The congestion of aircrafts at the track header of a certain RNW waiting to carry out its corresponding operation is avoided. The result will be given by the number of times the established limit is exceeded.

### 3.3.3 | Weighted factor

The principle behind constraint handling techniques (CHT) is how to combine the optimization process and the constraint satisfaction. There are two main ways to impose constraint handling: indirect constraint handling transforms constraint objectives into optimization objectives; while direct constraint handling explicitly enforces the fulfillment of constraints during the local search (Coello Coello, 1999; Liang et al., 2023). In this work, the selected CHT has been the indirect constraint handling.

The most popular CHT methods are the static penalty method (SPM) and the dynamic penalty method (DPM). Static penalty methods use fixed penalty coefficient values during the evolutionary process. The general expression for the static penalty function is presented in Equation 7.

$$F(x) = F_p(x) + R \cdot [\sum_{i=1}^{\#CE} p_i(x)] \tag{7}$$

where $R$ is a fixed value that determines the weight of the constraint satisfaction in the fitness function, $\#CE$ is the number of constraint equations, and $p_i(x)$ are the failures or penalties from the constraint equations.

On the other hand, in dynamic penalty function methods, the penalty parameters change with the evolutionary process. The general expression for the dynamic penalty function is presented in Equation 8.

$$F(x) = F_p(x) + (C \cdot t)^\alpha \cdot [\sum_{i=1}^{\#CE} p_i(x)^\beta] \tag{8}$$

where $C$ and $\alpha$ are user-defined constants, $\beta$ (usually valued 1 or 2) is another user-defined factor that powers the penalty values and $t$ is the number of the generation that is being evaluated. It is clear that as t increases, the penalty term increases. This means that in the beginning of the optimization less penalty is given to the optimal function, and the closer we get to the end, the less infeasible solutions are accepted.

There are different variations of the DPM. For this study, the annealing penalty method (Liang et al., 2023; Pereira & Fernandes, 2010) was selected. This method is based on the idea of simulated annealing. Equation 9 presents the expression for this method.

$$F(x) = F_p(x) \cdot [2 - e^{-\alpha}] \qquad \alpha = \sum_{j=1}^{\#CE} p_j(x)^\beta / T \tag{9}$$

As in simulated annealing, it is necessary to define the cooling scheme for the parameter $T$. From the different methods that exist, the following four are the most commonly used: Alpha (see Equation 10), Boltzmann (see Equation 11), Cauchy (see Equation 12) and Square Root (see Equation 13).

$$T_{Alpha} = T_0 \cdot 0.98^t \tag{10}$$

$$T_{Boltz} = \frac{T_0}{1 + \log t} \tag{11}$$

$$T_{Cauchy} = \frac{T_0}{1 + t} \tag{12}$$

$$T_{Sqrt} = \frac{T_0}{\sqrt{t}} \tag{13}$$

where $T_0$ is the initial temperature and $t$ is the generation of the evolutionary process being evaluated. Figure 4 shows the evolution of the four cooling schemes as the process evolves. Important to take into consideration that in some cases the value of the temperature tend to be zero very early loosing the capacity to work efficiently as a weighting parameter. In these cases, as the number of evaluations is a fixed parameter, it is recommended to increase the initial temperature $T_0$.

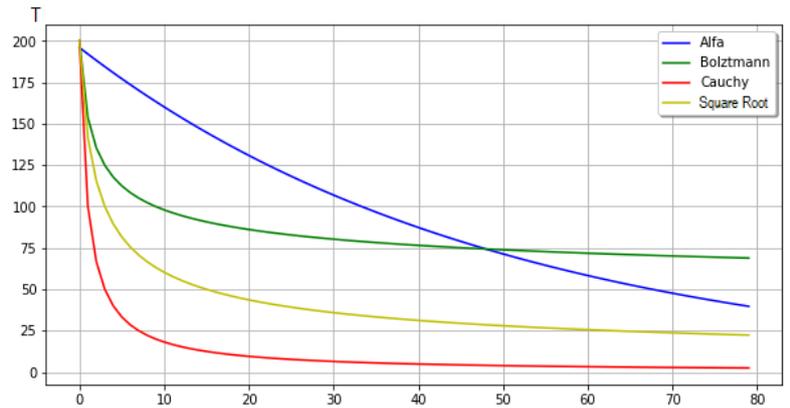

**FIGURE 4** Temperature variation according cooling scheme

### 3.3.4 | Total Fitness

Total Fitness is, ultimately, the value to be optimized by taking into account not only the Pure Fitness value but also the results obtained from evaluating the constraint equations for every individual of the population. As this is a minimization problem, that is sought to minimize the value of the total contamination of LTO airport operations, then the total fitness function is computed by adding the value of the Pure Fitness to the weighted sum of the different constraint penalties. Depending on the CHT selected, the weighted sum of constraint penalties will vary as explained in the previous subsection 3.3.3. In any case, each constraint penalty will have a weighting coefficient associated, as the BG constraints do not have the same importance of fulfillment as the RNW constraints have.

This process could be summarized by saying that the objective of the problem is the optimization of a multi-objective function (contamination, BG constraint, RNW constraint) in such a way that the weighting coefficients establish the levels of exchange between the different criteria.

Finally, it remains to determine how the weighting factors have to be established. In the case of DPM with simulated annealing, the only factor to determine is the value of the initial temperature $T_0$, which is usually set between 100 and 200. In the case of SPM, the weighting factors must be determined based on the level of exchange between pure fitness and the constraint penalty values. In our model, they have been set at 100 for the BG constraints and 50 for the RNW. This means that an individual with an additional BG error will be equivalent to another individual that has a Pure Fitness value 100 units (minutes-contamination) lower.

## 3.4 | Genetic processes

The evolutionary process of the GA has a perfectly defined work sequence, where the difference in adapting the procedure to a specific problem lies in the set of hyperparameters that are selected. The main GA procedures are:

**Elitism**: mechanism by which the best individual in the current population is selected and included directly as a member of the new evolved population in the next generation. It has been demonstrated (Rudolph, 1994) that the retention of the best individual in each generation is a necessary condition for the GA to converge to the optimum. A question arises as to what is meant by the best individual, the one with the best total fitness or the one with best pure fitness? The one that best meets the constraints imposed? All of them at once? In the present model, the one with the best Total Fitness has been selected as the best individual, since it is understood that this value includes, in one way or another, all the aptitudes of the individual against the proposed problem.

**Selection**: procedure by which the GA selects individuals, called parents, from the current population to reproduce in children who will become part of the new generation. The most common selection techniques are roulette-wheel selection and tournament selection. In this case, tournament selection has been chosen. For each tournament, *T* individuals are randomly selected from which the one whose fitness function gives the best rating is selected for reproduction, transferring its genetic load to the corresponding children. To avoid a very high selection pressure on the individuals with the best fitness, a parameter is introduced that, based on a random flag, establishes whether the selection is for the best or for the worst of the selected individuals.

**Crossover**: this reproduction process is carried out by selecting two parent chromosomes and generating new chromosomes. The most common crossover operators for integer array representations consist of aligning and dividing the parent chromosomes into parts to finally exchange the fragments with each other, giving rise to the children that will make up the new population. These crossover operators include uniform crossover, where each gene in the child chromosome is randomly selected from one of the parents with equal probability; and N-point crossover, where N gene positions from the chromosome are randomly selected and each chromosome subsection formed from the cuts at these genes is selected from a different parent (e.g. in two-point crossover, three subsections are generated, where the corner subsections are taken from one parent while the center one is taken from another). In the LTO problem, as each gene represents an ATM operation assignment, these crossover operations will generate new individuals from the combination of ATM assignments from both parents. Figure 5 presents an example of 2-point crossover for part of the chromosomes.

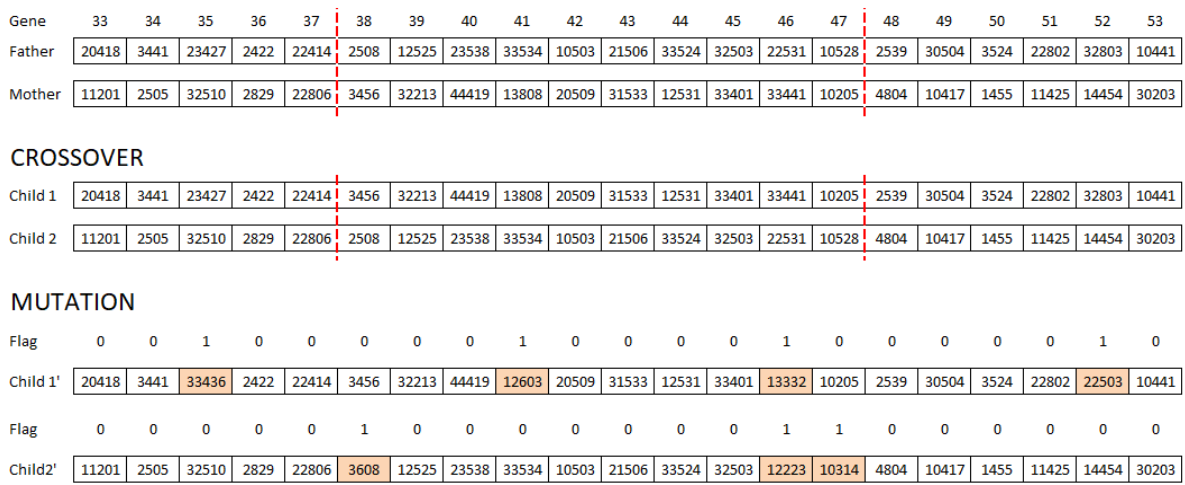

**FIGURE 5** Crossover and Mutation examples

**Mutation**: consists in the random alteration of a few genes of the children. Its purpose is to introduce randomness into the generational structure of the new population. It will be applied based on a parameter that establishes the threshold from which said procedure will be carried out or not. A factor that must be determined is the possible evolution of the mutation threshold. There are several trends: keeping it constant, increasing, or decreasing throughout the evolutionary process. In case of applying a variable mutation coefficient, it is not convenient to make the modification in each and every one of the iterations of the process in order to give time for changes made in the mutation level to have an effect on the evolution of the population. Emphasize the importance of maintaining the chromosomal structure of the child during the mutation process. It must be guaranteed that the values of the mutated genes meet the constraints imposed by the variation intervals of the alleles. In order to ensure this in the LTO problem, the same process used with the generation of the initial population for each gene is employed in the mutation process. In Figure 5, an example of uniform mutation for part of the chromosomes is presented.

**Replacement**: To build the new population for the next generation, the most common method is generational replacement, where parents are replaced by offspring, regardless of whether fitness is improved or not. This non-deterministic and non-overlapping strategy increases the generational gap between the new and the old population. In order to avoid losing good individuals, this replacement strategy is used in combination with elitism, so the best individual remains the guide of the convergence process. Another possible strategy proposed by Wu et al. (Wu, Liu, & Peng, 2014) is the use of parent-offspring replacement through competition (Best Parent-Child Replacement). For this purpose, once the two offspring for two parents have been obtained, the two individuals with the best fitness values, either parents or offspring, are selected as members of the new population. This way of working guarantees the inclusion of the best individual in each generation, so it is not necessary to activate Elitism.

## 4 | EXPERIMENTATION

## 4.1 | Experimental setup

The evaluated problem considered in this work corresponds to the real-case flight data from New York's JFK airport corresponding to August 5, 2019. The airport scheme can be seen in Figure 6.

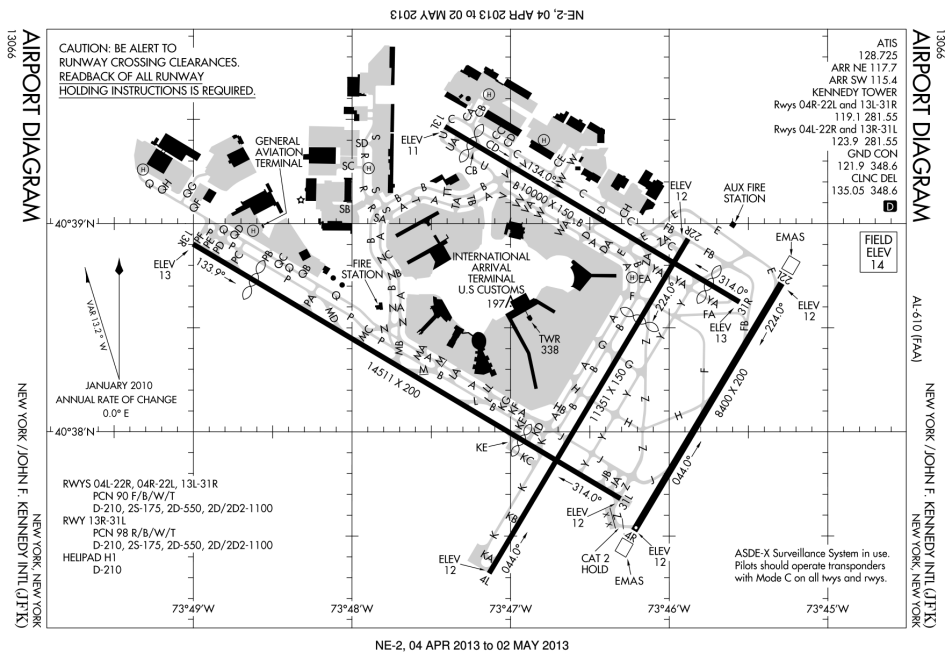

**FIGURE 6** JKF Airport diagram

**Airport Configuration**: JFK airport consists of 6 working TERminales and a total of 168 BGs (126 gates and 42 remote stands), whose distribution can be seen in Table 2.

**TABLE 2** Terminals and BGs (Gates/Remote) at JFK airport

| Terminal | Description | BGs |
|---|---|---|
| Terminal 1 | Together with Terminal 4 is able to handle Airbus A380 superjumbo aircraft. | 11 / 3 |
| Terminal 2 | It is only used by Delta Airlines. | 10 / 5 |
| Terminal 4 | It handles international arrivals. It serves as a major international hub for Delta Airlines for its long-haul flights and other airlines. | 38 / 20 |
| Terminal 5 | Operates with various airlines but JetBlue Airways is the major one. | 26 / 14 |
| Terminal 7 | Currently operated by British Airways but there are more airlines operating there as well. | 12 / 0 |
| Terminal 8 | The largest passenger terminal at JFK Airport. It is able to handle nearly 13M passengers annually. | 29 / 0 |

On the other hand, JFK airport has 4 RNWs, whose configuration is presented in Table 3.

**ATMs scheduled**: An analysis of the initial information of the ATMs was carried out by cleaning the data from outliers in order to obtain a final coherent data set. The ATMs where any information about the terminal, airplane or schedule time was missing were eliminated. One of the data that has not been provided in the list of real ATMs is the BG that was assigned. The lack of such information has made it impossible to compare the pollution results obtained through the GA model with those that would have really occurred, making it impossible to obtain the improvement that the application of the GA model results would entail in the pollution of the airport's LTO operations.

As a result of the clean process, the final general parameters of the problem to be solved were:

**TABLE 3** RNWs at JFK airport

| RNW | Direction | Length (m) | Surface |
|---|---|---|---|
| 1 | 13L/31R | 3048 | Concrete |
| 2 | 13R/31L | 4423 | Concrete |
| 3 | 4L/22R | 3682 | Concrete |
| 4 | 4R/22L | 2560 | Asphalt |

- Initial number of movements - 1,309
- Number of movements after cleaning outliers - 1,247
- Number of final ATMs - 721

  · ATMs with LAN/TOF - 526
  · ATMs with only LAN - 101
  · ATMs with only TOF - 94

An example of some of the scheduled flights is presented in Table 4.

**TABLE 4** Example of flights schedule

| Flight | LAN | TOF | Airline | Aircraft | Orig/Dest |
|---|---|---|---|---|---|
| AV670 | 02:23 | 04:31 | Avianca CR | A319neo | SAL/SAL |
| IB2627 | 22:17 | 23:46 | Iberia | A330-200 | BCN/BCN |
| LY1 | 04:59 | N/A | El Al Israel | B777-200 | TLV |
| LY7 | 17:45 | 20:06 | El Al Israel | B787-9 | TLV/TLV |
| KQ2 | 08:03 | 13:16 | Kenya Airways | B787-8 | NBO/NBO |
| KU117 | 15:50 | N/A | Kuwait Airways | B777-300ER | KWI |
| SQ26 | 10:38 | 20:46 | Singapore Air. | A380-800 | FRA/FRA |
| BW16 | N/A | 02:10 | Caribbean Air. | B737-800 W | /KIN |
| EY101 | 16:25 | 22:55 | Etihad Airways | A380-800 | AUH/AUH |
| EK201 | 14:41 | 22:59 | Emirates | A380-800 | DXB/DXB |

Before starting with the resolution of the model, the capacity of the airport to attend the scheduled operations was evaluated. For this, the ATMs were accumulated according to their temporal sequence and the programmed terminal. As can be seen in Figure 7, the airport configuration is compatible with the daily scheduled ATMs. Terminals 1, 2, 4 and 5 present peaks of activity that almost suppose the total use of the available BGs.

**Airplanes Contamination Level**: According with the final list of ATMs, as much as 40 different types of airplanes were used. To evaluate the contamination that each of those different aircraft produce during the LTO operations, we have used the file "1.A.3.a Aviation - Annex 5 - LTO emissions calculator 2019-2020.xlsm"[§] from (Winther & Rypdal, 2020), which splits the contamination values in 3 groups:

- Mass of fuel burnt (kg)
- Mass of CO, HC, NOX, CO2 emitted (kg)
- Mass of total Particulate Matter (PM) emitted (kg)

The contamination values are normalized to subsequently obtain the pollution factor for each aircraft by weighting the individual values for each of the series. We have selected the following weighting values: 0.85 for the mass of fuel burnt; 1.25 for mass of gases emitted; 1.15 for mass of total PM.

In Table 5, together with the computed pollution factor, appears an additional factor named Typology. This factor characterizes the airplane from the point of view of which of the RNWs it can use for LAN and TOF operations.

---

[§] https://www.eea.europa.eu/publications/emep-eea-guidebook-2019/part-b-sectoral-guidance-chapters/1-energy/1-a-combustion/1-a-3-a-aviation-1-annex5-LTO/view

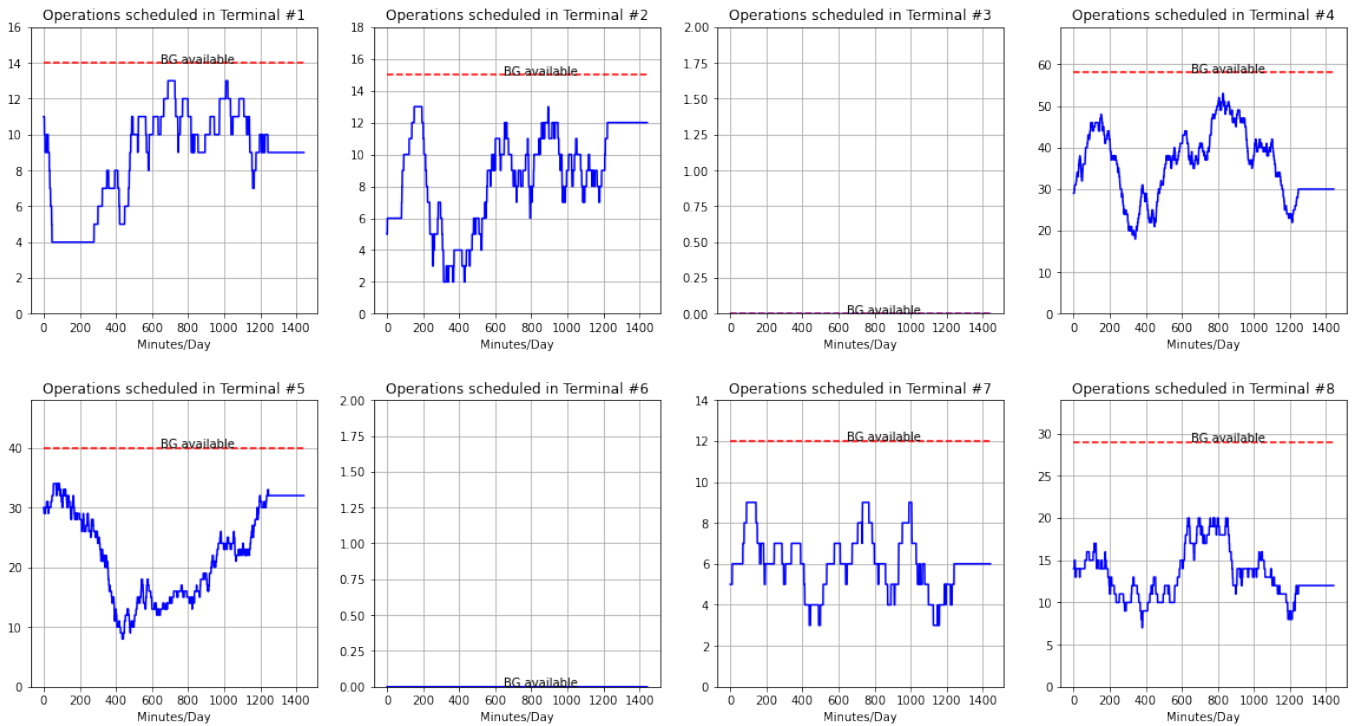

**FIGURE 7** Terminal occupation in JFK airport

**TABLE 5** Example of airplanes contamination factor

| Aircraft Model | Passangers | Pollution factor | Typology |
| --- | --- | --- | --- |
| 140 | 44 | 1.0000 | 1 |
| A320-200 | 180 | 3.9627 | 2 |
| A330-300 | 300 | 6.5764 | 2 |
| A340-300 | 335 | 10.5235 | 3 |
| A350-900 | 315 | 6.4813 | 3 |
| A380-800 | 615 | 11.7671 | 3 |
| B737-300 | 148 | 2.4314 | 1 |
| B737-900ER Winglets | 206 | 2.8659 | 2 |
| B747-8i | 352 | 9.8865 | 3 |
| B767-300 (winglets) | 261 | 5.6577 | 2 |
| B777-300ER | 378 | 9.1470 | 3 |
| CRJ200 LR | 50 | 1.2143 | 1 |

- Typology 1 - RNW01 (50%) - RNW04 (50%)
- Typology 2 - RNW01 (25%) - RNW02 (25%) - RNW03 (50%)
- Typology 3 - RNW02 (100%)

**Hyperparameters of the model**: Within the large number of combinations available, the hyperparameters presented in Table 6 have been selected for the resolution of the Base Model.

The population size value (150) ensures a population large enough to, guaranteeing population diversity, solve the problem with the minimum size significantly reducing the resolution time when working with large airports. We also tested populations of 100 individuals, but the diversity was not high enough; and 200 individuals, but the results were similar to those obtained with 150.

The number of generations (1.500) has been established taking into account that, after a few resolutions of the model, 400 is approximately the average number of generations necessary to obtain a problem with zero errors in both the BG assignment and the RNW assignment, giving another 1100 generations to the model to improve the value of pure fitness. From the experiments performed with the base model, it was shown that the improvement after this number of generations was very little. Also, in a pre-experiment, we could check that for other airport configurations of smaller complexity, the number of generations needed to converge is smaller than this value.

**TABLE 6** Hyperparameters of the Base Model

| Hyperparameter | Value |
| --- | --- |
| Population size | 150 |
| Number of generations | 1.500 |
| Max ATM to BG | 10 |
| Max ATM to RNW | 7 |
| Penalty Method | SPM |
|     Pure Fitness | 1 |
|     BG errors | 100 |
|     RNW errors | 50 |
| Elitism | Not Active |
| Selection | Tournament selection |
| Tournament size | 2 |
| Probability of selecting worst | 20% |
| Crossover | One-point crossover |
| Crossover probability | 100% |
| Mutation | Uniform mutation |
| Mutation probability | Decreasing with initial value 0.60% and final value 0.10% |
| Replacement strategy | Best Parent-Child Replacement |

The SPM presents weighting values of (1, 100, 50). As the variation of pure fitness during the model resolution is roughly 8.000 units, the BG penalty coefficient (100) implies that one BG error weights as much as 1.25% improvement in the pure fitness. In the RNW case, the interchange cost goes down to 0.625%. These values force the model to search intensely for the disappearance of assignment errors before proceeding with the improvement of pure fitness. In section 4.3.3 we analyze the performance of different constraint handling techniques for this problem.

A tournament selection was chosen instead of roulette wheel selection in order to give higher chances to individuals with lower fitness values. Through the study of different values of $T$ number of competitors, it was proven that the best value was $T = 2$ competitors. Due to the replacement strategy used, since the process is highly inbred, the percentage that allows the worst competing parents to be selected instead of the best is raised to 20% in order to increase population diversity (considering that with generational replacement with elitism a probability of 5% was enough).

The one-point crossover is the one that has presented the best genetic behavior within the different alternatives. The use of two-point or uniform crossover showed that the effect of the mutation process altering genetic convergence is amplified. The crossover probability was established at 100% as the replacement strategy asures to maintain the best individuals from generation to generation.

In this study, a uniform mutation with a decreasing probability strategy was used. The levels have been established between 0.60% and 0.10% which means, considering a 120-chromosome length, mutating between 4.5 and 0.6 alleles for each chromosome of each individual in the evaluated population. In section 4.3.2 we analyze the performance of different mutation criteria for this problem.

Finally, both generational replacement using elitism and best parent-child replacement strategies were compared. As can be seen in the Figure 8, the behavior of the Best Parent-Child replacement strategy produces better results. The constraint equations are satisfied in fewer iterations, 65 vs 220, and the total fitness optimization achieves better values with fewer generations (3,0% improvement in total fitness with a 40% reduction in the number of iterations). In view of the results, the following study will use the best parent-child replacement strategy.

## 4.2 | Experimental results from the base model

In this section, we evaluate the results obtained from the base model, and how the fitness values evolved. Of a total of 31 executions, the median was selected for this study.

### 4.2.1 | Population evolution

To evaluate the behavior of the population evolution, the mean values of total fitness are used as the best indicator. As can be seen in Figure 9, as the genetic process evolves, both the mean and worst values of population fitness tend to converge very soon at the best fitness of each generation. The same effect can be seen by analyzing the standard deviation of the population total fitness. Given the inbreeding of the chosen process, the best and worst total fitness values tend to be very close as early as the 300th generation. This does not mean that there are no differences between individuals in the population, but that these differences are concentrated in a small number of alleles.

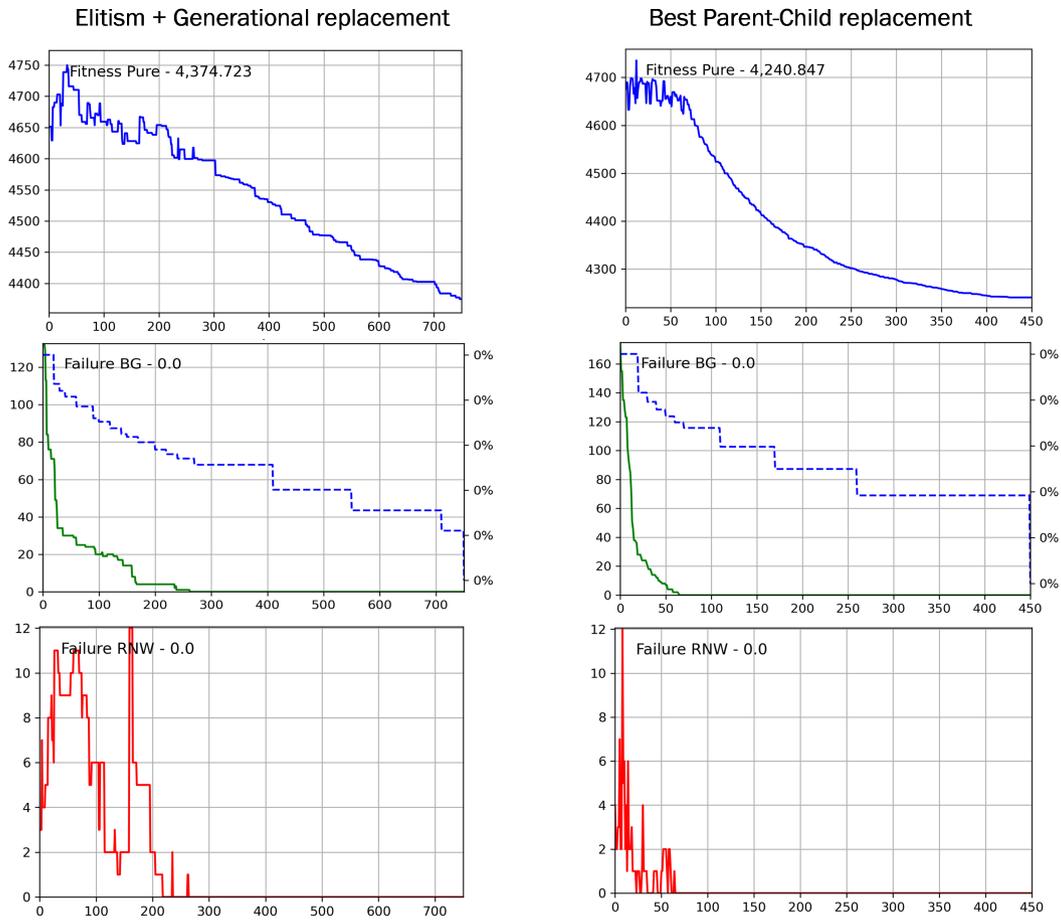

**FIGURE 8** Comparative Replacement criteria

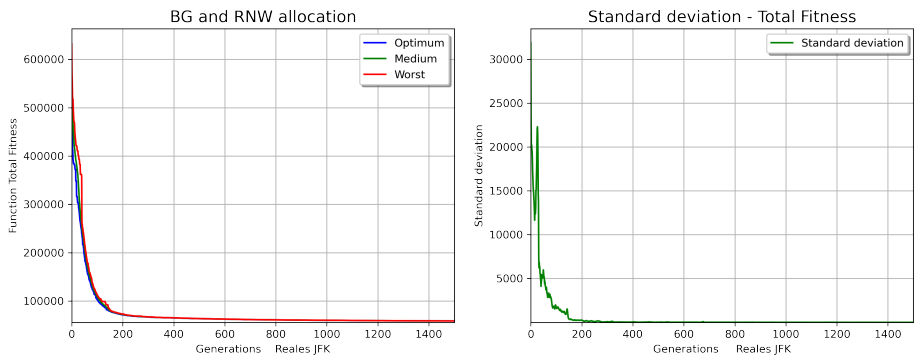

**FIGURE 9** Population evolution

As a summary of the above, both the convergent evolution of the population and the optimisation process of the different components of the total fitness can be clearly seen in the Figure 10. The condition for assigning LAN/TOF operations to the different RNWs as well as to the BGs is solved for almost all individuals. During the first 50 generations the GA focuses its attention on solving the constraint equations keeping the Pure Fitness unchanged. Due to the large inbreeding of the process, in generation 50 the PF values start to show less dispersion than in the initial population. From generation 50 until the end of the evolutionary process, the GA only improves the PF value.

In relation to the assignment to the BGs of the terminals, the errors present in the final population are very small with a mean value close to 125 from the 4.550 errors present in the initial population. Finally, the PF presents an ideal evolution with an important reduction value over the initial one.

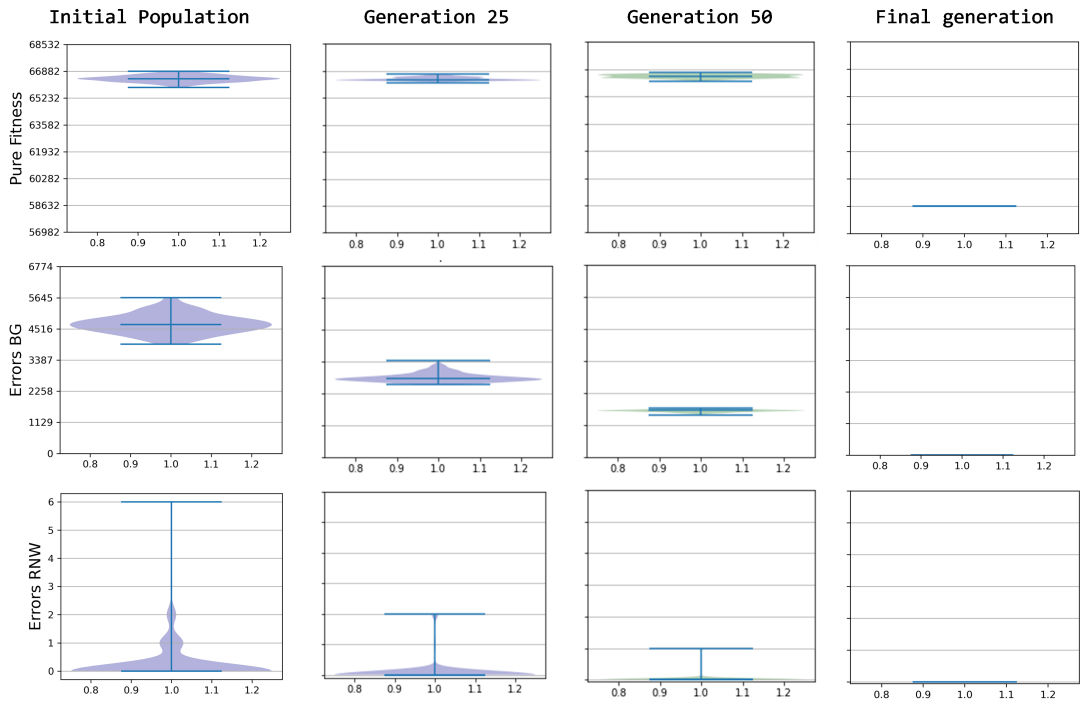

**FIGURE 10** Population evolution

## 4.2.2 | Fitness evolution

The total fitness adjustment process carried out by the GA presents a similar behavior in almost all the scenarios studied, initially adjusting the BG/RNW penalty and then starting with the fine adjustment of pure fitness.

Figure 11 shows the evolution of the three components of total fitness, both for the average fitness values of the population and for the best individual (optimal solution) in each generation. Practically in generation 200 an almost perfect compliance with the sequence of assignments of the ATMs to the BGs/RNW has been achieved at the cost of sacrificing the pure fitness value. From said generation to the maximum number of generations, established in 1.500, the GA seeks to improve pure fitness.

In the same figure, the graphic 'BG Failure - Average Values' shows the behavior of the mutation coefficient. Every 10 generations, the total fitness of the population is evaluated and in the event of a significant improvement, the coefficient is adjusted by steps reducing its value until reaching the minimum mutation value in the last generation. It can be clearly seen how the mutation coefficient follows the evolution of the Pure Fitness graph due to the importance of said value in the total fitness. This is due to the way the selected mutation criteria (decreasing) works. At the beginning of the evolutionary process, the selected value is 0.60%, which means that 1 for every 150 genes suffers mutation. As the process evolves, the mutation factor decreases until reaching the final value of 0.10% which implies 1 mutation for every 1.000 genes.

On the other hand, the comparative evaluation between the best individuals in each generation with the precedent one allows us to visualize the changes made in the values of the genes. In order to study the evolution of the composition of the best individual and to be able to evaluate how the GA has been searching for the local optimal solution, Figure 12 was generated.. It can be seen that during the first 500 generations very significant changes occur in the structure of the optimal chromosome. From then on, punctual modifications occur where only few of the total genes are modified. In the last generation the final reconstruction process has no effect, as the model had previously reached a solution with zero errors.

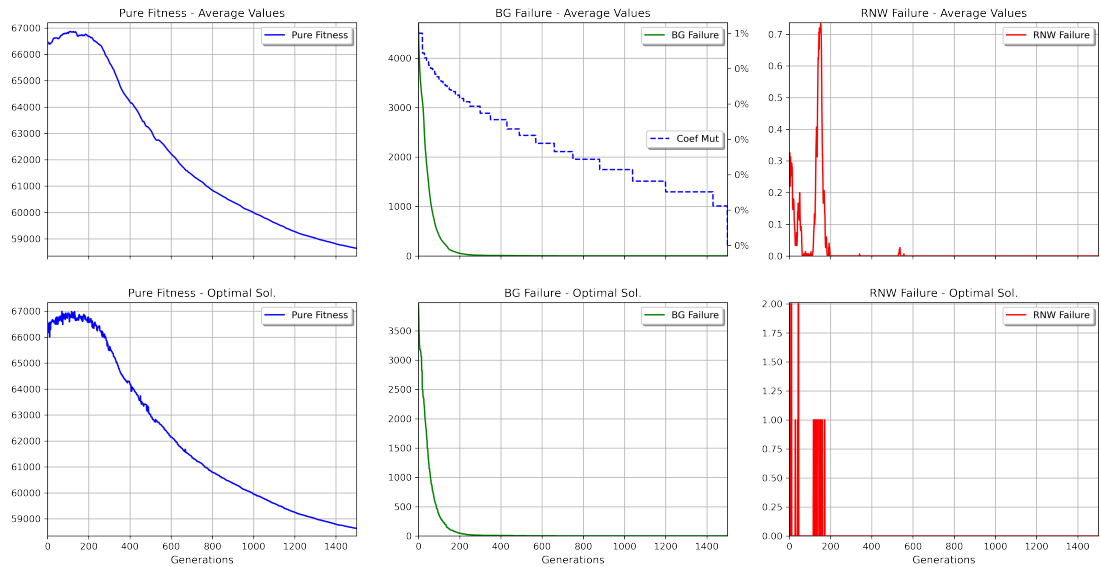

**FIGURE 11** Fitness evolution of the average and best individual in each generation

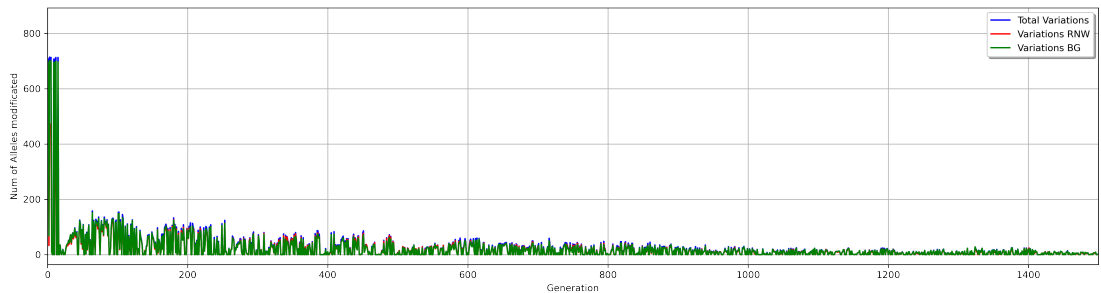

**FIGURE 12** Genes modification of best individual during evolution of GA

### 4.2.3 | Analysis of the solutions

In the previous section has been shown that the GA procedure is a good method for solving the problem of assigning BGs and RNWs to ATMs subject to constraints.

There are four principal factors that guide the assignment process: the schedule of the flight, the level of occupancy of a specific BG, the distance of the BG to the RNW head and, finally, the level of contamination of the airplane. Figure 13 resume the number of ATMs assigned to the BGs ensuring that the number never exceeds the threshold fixed in 10. Although none of the BGs reaches the established limit, in Terminal 2 BGs 10 and 14 scheduled 9 daily movements.

Furthermore, Figure 14 shows the six ATMs that have been assigned to the BG number 8 of Terminal 1. In this special case, the initial ATM as well as the final one only have one movement, meaning that the BG was occupied at the beginning of the studied period and will also be occupied at the end of that period of time. According to the flight schedule assigned to this BG (Table 7), it is not occupied for just 10% of the 24 hours.

Finally, Figure 15 shows the sequence of LTO operations that are carried out through RNW 5, clearly showing the high occupation of this specific airport infrastructure. In Figure 16, the assignment of RNWs to ATMs assigned to terminal 4 can be seen, which is the one with most ATMs.

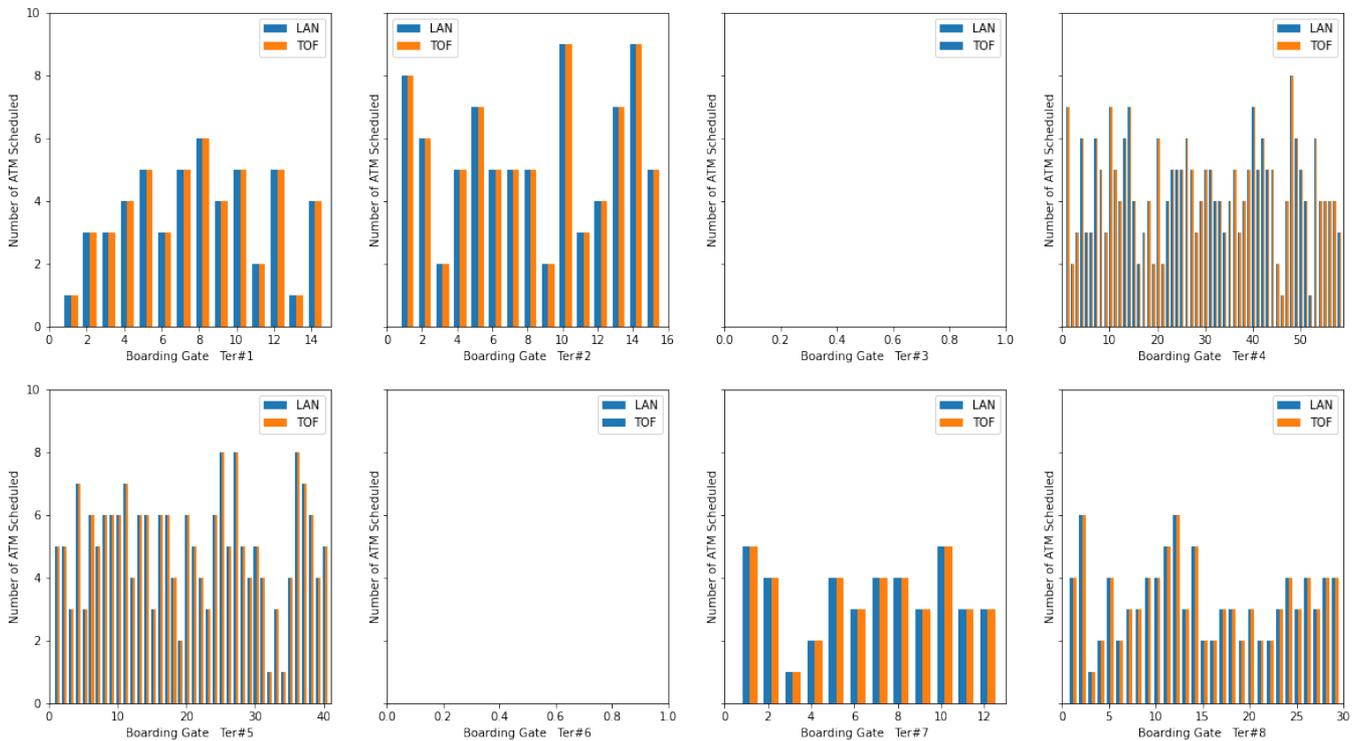

**FIGURE 13** ATMs by BG and by Terminal

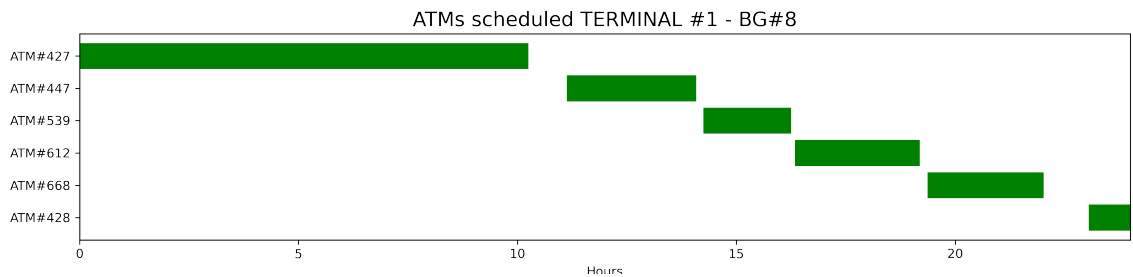

**FIGURE 14** ATMs assigned to BG#08 of Terminal#01

## 4.3 | Sensitivity Study

Previously, we have been analyzing the results of the Base Model subject to the parameters of SPM and decreasing mutation. In this section, we compare the results of the GA model subject to variations in the mutation criteria as well as variations in the CHT used. For each of the experiments, 31 executions were generated for each of the scenarios, storing the values of time processing, Pure Fitness, errors in BG and errors in RNW.

Before starting the sensitivity analysis, we consider necessary to establish the level of optimization achieved with the use of GA. A comparison has been made with the results obtained by applying a binary Linear Programming Model.

Once the optimal reference of the problem is obtained, we present the different behavior that the model presents subject to modification in both the mutation criteria and the penalty method. Later on, we expose normality and homogeneity studies of the model, and some discussions on the results as well as a method to select the best CHT criteria.

### 4.3.1 | Comparison with a Linear Model

In order to establish the level of optimization achieved with the use of the GA, a binary Linear Programming Model has been applied to the LTO optimization problem to comapre the results. The commercial framework LINGO (Schrage, 1999) was used for this purpose.

**TABLE 7** Flights scheduled at BG#08 of Terminal#01

| Flight | LTO | Time | RNW | Airline | Orig/Dest | Aircraft |
|---|---|---|---|---|---|---|
| KX793 | TOF | 10.25 | 1 | Cayman Airways | GCM | B737-300 |
| KE81 | LAN | 11.13 | 2 | Korean Air Lines | ICN | A380-800 |
| KE82 | TOF | 14.08 | 2 | Korean Air Lines | ICN | A380-800 |
| MU587 | LAN | 14.25 | 2 | China Eastern | PVG | B777-300ER |
| MU588 | TOF | 16.25 | 2 | China Eastern | PVG | B777-300ER |
| AF6 | LAN | 16.34 | 2 | Air France | CDG | A380-800 |
| AF7 | TOF | 19.19 | 2 | Air France | CDG | A380-800 |
| AF10 | LAN | 19.37 | 2 | Air France | CDG | A380-800 |
| AF11 | TOF | 22.02 | 2 | Air France | CDG | A380-800 |
| KX792 | LAN | 23.05 | 1 | Cayman Airways | GCM | B737-300 |

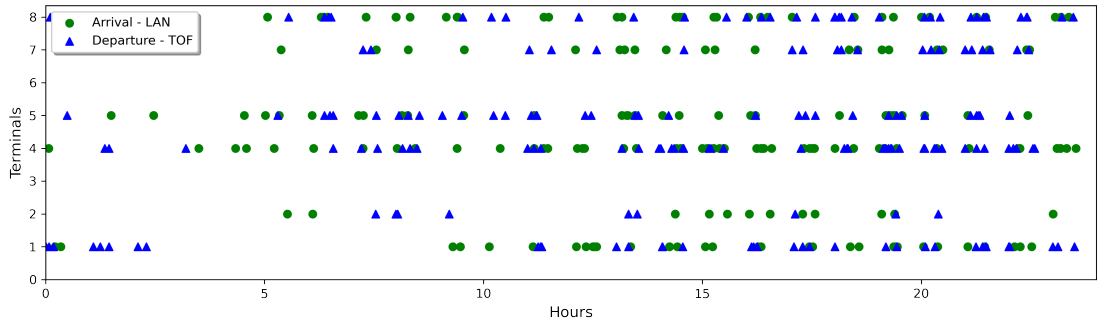

**FIGURE 15** LTO operations assigned to RNW#02

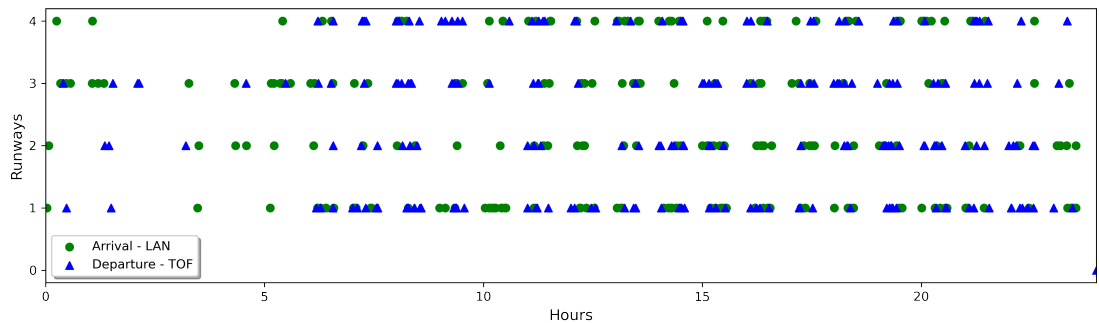

**FIGURE 16** RNW assigned to LTO operations from Terminal#04

The biggest handicap of working with binary Linear Models is the immense amount of binary variables they require. If we tried to model the problem previously evaluated, we would need to have a software capable of handling up to a quarter of a million of binary variables. In our case we only have a program that is exclusively capable of handling a maximum of 3,200 binary variables. This is also due to linear models being incapable of dealing with the scalability of NP-hard problems. Because of this, it has been necessary to reduce the size of the model evaluated.

The new model that will be used from now on, has been generated by taking a section of the previous model. In total, 120 ATMs have been selected, redefining the airport configuration to 2 Terminals with 7/6 BG respectively and 2 RNW. To minimize the number of binary variables, each ATM has been assigned exclusively to one terminal.

The MILP model that solves the new problem including the constraint equations requires 3,100 binary variables and 3,008 constraint inequalities to solve the objective function. By using the 'Branch and Bound' method, an optimal value of the fitness function of 4,194.84 minutes-contamination is obtained.

The hyperparameters used in the GA model are: 150 population size; 600 number of generations; decreasing mutation coefficient [0.50%,0.15%]; Static Penalty Method with values [1,100,50]. As can be seen in Figure 17 the model reaches around the generation 400 the Total Fitness optimal value of 4.217,94 minutes-contamination which is only 23,10 minutes-contamination (0.55%) higher than MILP result.

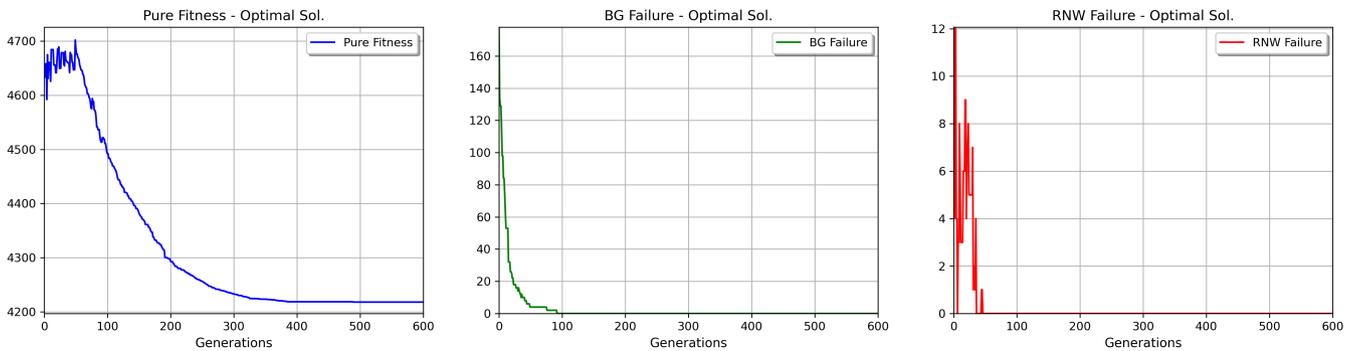

**FIGURE 17** Fitness evolution of the best individual in each generation (120 ATMs)

### 4.3.2 | Mutation criteria

One of the hyperparameters that must be carefully established is the mutation criteria due to the important effect that it has in the problem resolution. The first criteria to determine is the level of mutation that makes the GA work correctly. Inappropriate values can cause that the GA convergence process disappear, turning the calculation process into a purely random one.

To establish the optimal level, the model has been evaluated with 20 different decreasing mutation levels, varying their maximum levels between 40.0% and 0.58% and with a constant minimum level of 0.15%. The Figure 18 clearly shows how high initial mutation levels 40.0% do not produce a convergence of the population towards optimal solutions. Mutation values around 8.0% beginning to show the convergence. With initial levels close to 0.50%, the convergence of the population towards the optimal solution is smooth and constant.

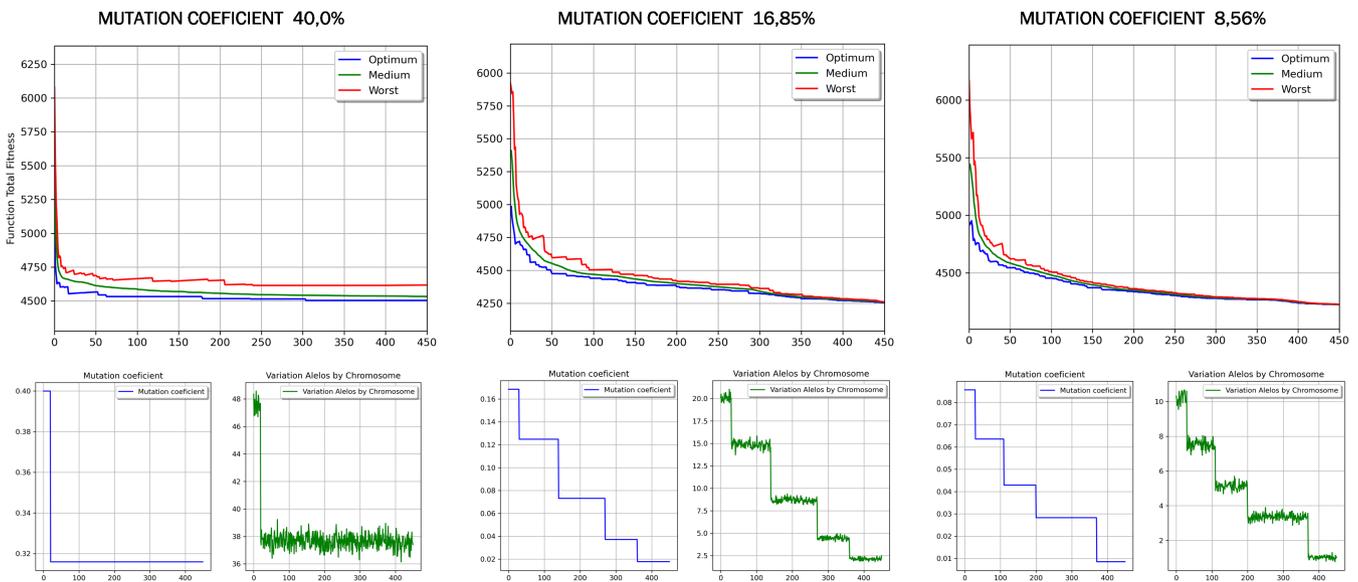

**FIGURE 18** Evolution of Population Fitness depending on the initial mutation value (120 ATMs)

Being important the behavior of the population, we must see how this initial mutation value translates into optimal fitness behavior. In the Figure 19 we can see that the Pure Fitness evolution improves as the initial mutation value decreases. In line with what has already been stated in relation to the convergence of the evolved population, the best behavior and final value of Pure Fitness is achieved with the low values evaluated of the mutation coefficient.

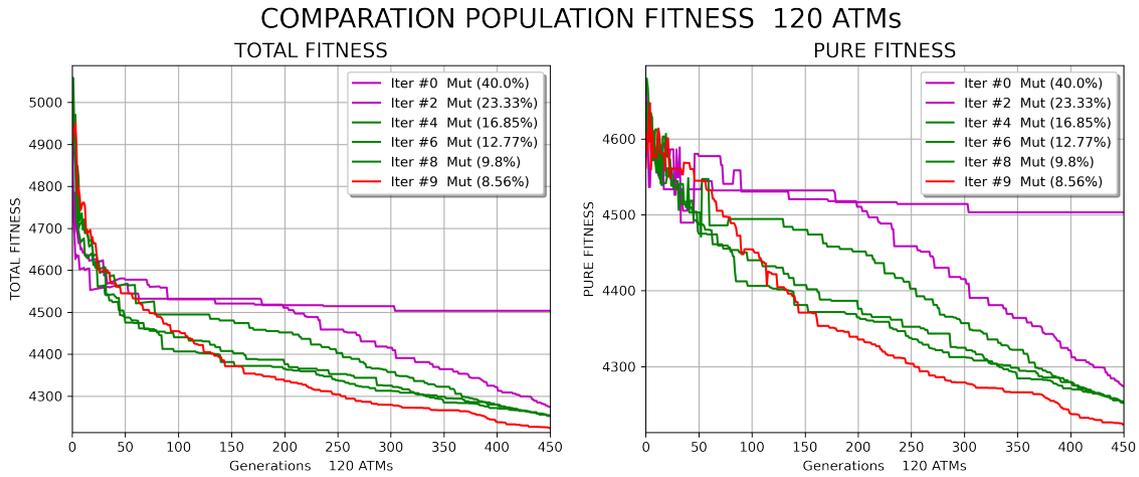

**FIGURE 19** Evolution of optimal Pure Fitness depending on the initial mutation value (120 ATMs)

Taken into consideration the above solution, in the rest of our study we have assumed the mutation interval $[0.10\%, 0.50\%]$ and, in order to reduce the calculation time, a number of generations of 450. There are three different criteria when applying mutation: decreasing (from 0.50% at the beginning of the evolution process until 0.10% at the end), constant (fixed in 0.30%) and increasing (from an initial 0.10% until 0.50%). In all of these models, the SPM was used as CHT.

In Figure 20 we present how the results of the Pure Fitness evolve depending on the mutation criteria selected. Although the decreasing mutation reaches the worst value 4.240,85 contamination-minutes at the end of the process, the three criterion present very close final values. The increasing mutation require more generations to achieve the complete fulfillment of the CE which indicates that higher initial mutation values help to find sooner the zero errors in the CE.

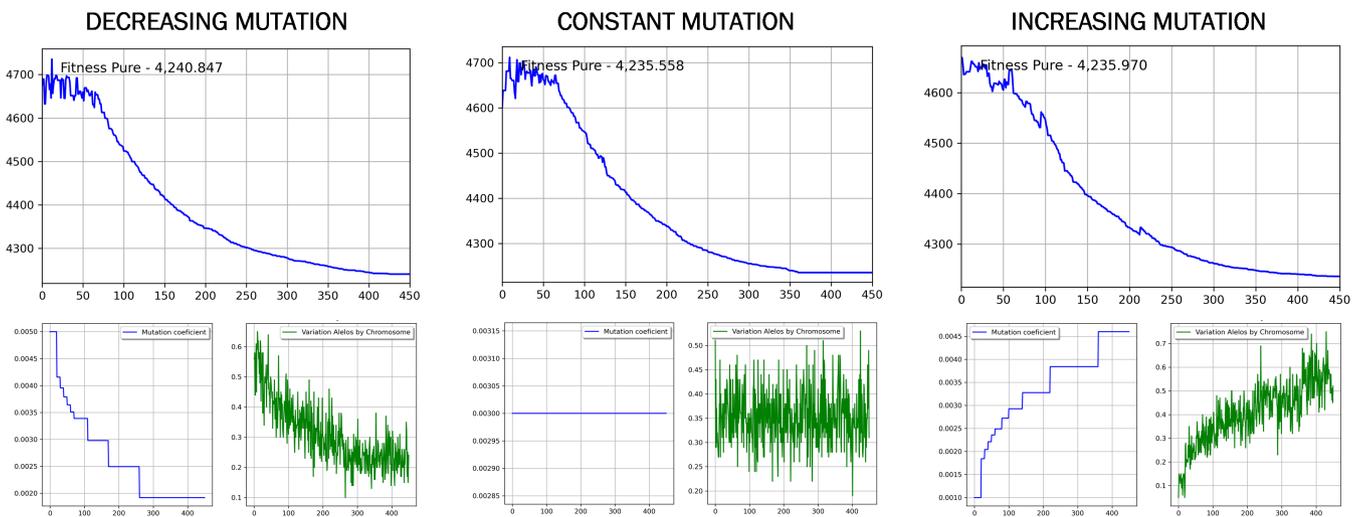

**FIGURE 20** Comparative of the Pure Fitness depending of the mutation criteria

Finally, Figure 21 shows the boxplots of the values obtained over the 31 executions for each of the different mutation criteria. There are not big differences between then. Only be noted that the behavior of the series of models corresponding to Decreasing Mutation shows greater dispersion than the rest of the criteria. Finally, it should be noted that the average time taken to calculate the different models was around 9.0 minutes.

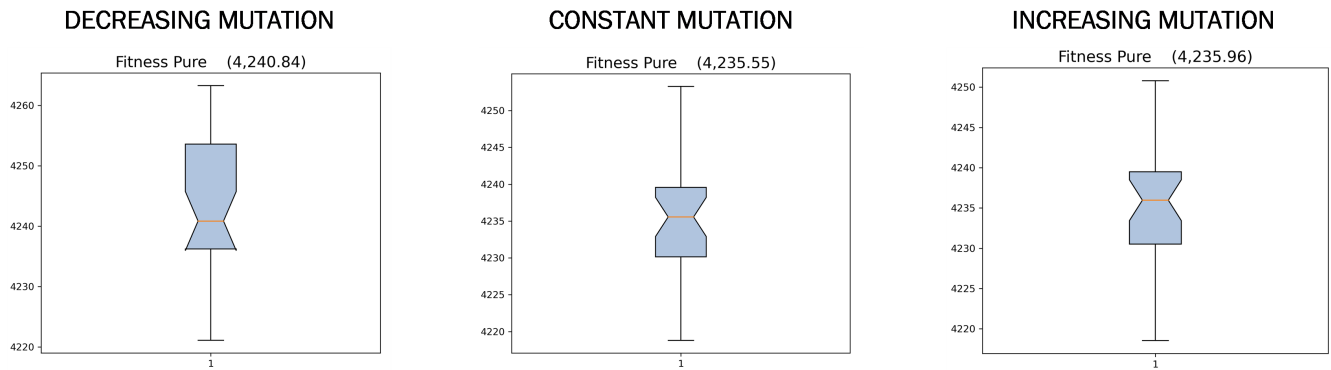

**FIGURE 21** Boxplots of fitness values for different mutation criteria

### 4.3.3 | Constraint Handling Methods

To evaluate the behavior of the model in the case of applying a DPM under the same assumption of mutation (decreasing), the annealing penalty method presented in Section 3.3.3 has been used, comparing the different temperature cooling schemes.

Figure 22 shows the evolution of the penalty factor according to the selected cooling scheme. The Alpha and Boltzmann schemes show similar behavior with a smooth and permanent trend of improvement as the evolutionary process progresses. Cauchy and Square root produce a similar behavior of the penalty factor where there are significant peaks. These peaks correspond to evolutionary sections where the errors remain constant (there are no improvements). As the evolution progresses and according to the cooling scheme, the temperature decreases, increasing the penalty term, i.e. as evolution progresses, less infeasible solutions are accepted.

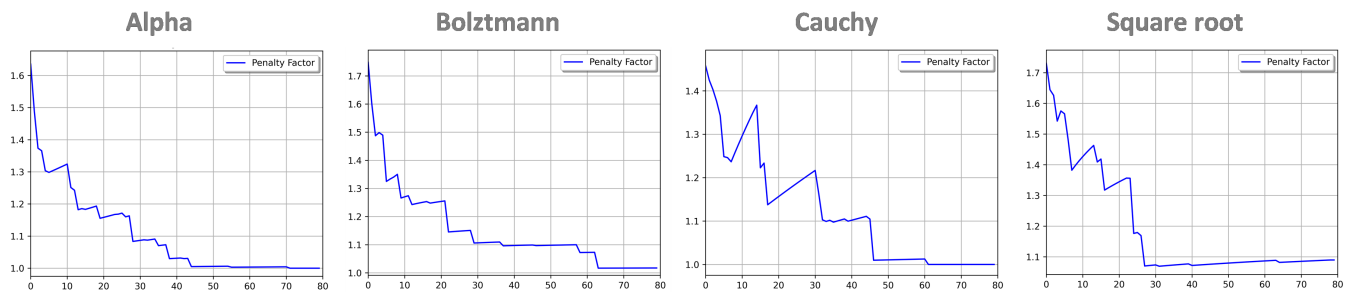

**FIGURE 22** Evolution of the penalty factor

In Figure 23 we can see the boxplots of the values obtained over the 31 executions corresponding to the DPM models. The runtime is similar between the different DPM models, approx 9,0 minutes, so the runtime graphs have been omitted. Although there are no big differences in behavior between the models, Alpha criteria is the most aggressive looking for the CE compliment while Bolztmann looks to be the most soft and constant. The average value of the different series evaluated show small differences between them, reaching, in all cases and taking into account the number of generations executed, very acceptable values with differences of 0.85% over the optimum value obtained by means of the linear binary model.

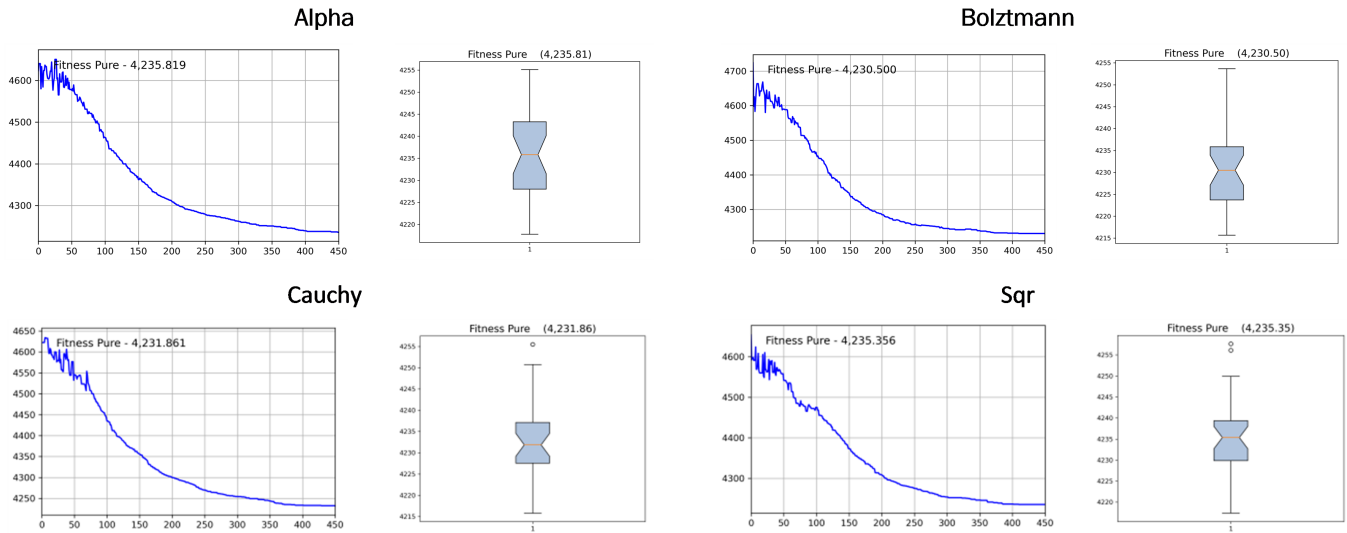

**FIGURE 23** Boxplots of fitness values for different DPM

## 4.3.4 | Contrast tests

In this section, contrast tests are used to study whether the models solved with different hyperparameters are the same. The normality of the Pure Fitness is evaluated using kurtosis and skewness, as well as the Shapiro-Wilk test and D'Agostino's K-squared test. The homogeneity is then studied through evaluation of the p-value, depending on the normality of the values compared, of a t-test or a Mann-Whitney-Wilcoxon U-test. Total Fitness is not evaluated as well-parameterized models with sufficient number of generations will obtain a final result with zero errors in both BG and RNW making the TF equal to the PF.

**Normality study**: The fact of not being able to assume normality mainly influences the parametric hypothesis tests (t-test, anova,...) and the regression models. In Figures A1-A7, the graphics for all the models are included, both the histogram and the theoretical normal curve.

In order to evaluate the normality of the series, two different methods are applied. The first was to evaluate the kurtosis and skewness of the theoretical normal curve. Both values, as seen in Table B1, confirm the normality of all the series. We also study the normality by analyzing the p-value of both the Shapiro and the D'Agostino tests. Shapiro test assumes as null hypothesis $H_0$ the normality of the series being rejected if the test value is close to zero. D'Agostino test aims to establish whether or not the given sample comes from a normally distributed population detecting deviations from normality due to either skewness or kurtosis. None of the tests, see Table B2, rejected the normality of any of the series evaluated.

**Homogeneity study**: Once the normality of every model has been guaranteed, the next step is to evaluate whether or not the different sets resolve the same problem using either the t-test or the Mann-Whitney-Wilcoxon U-test. To evaluate the behavior of the models subject to the SPM, the t-test has been used when having a series of normal values checking if the means of two normally distributed populations are equal. The hypothesis $H_0$ is accepted only when comparing models with increasing and constant mutation (see Table B3). The Mann-Whitney-Wilcoxon U-test reinforce the previous results concluding that increasing and constant series have the same variance and that both generate equidistributed set of results. From this it can be concluded that while increasing and constant mutation resolve the same problem, decreasing mutation is statistically different.

For the analysis of the DPM models, the Mann-Whitney-Wilcoxon U-test was used. The null hypothesis is that given two values x and y randomly obtained from two independent samples, the probability that x is greater than y is equal to the probability that y is greater than x. Therefore, the test contrasts whether two samples come from equidistributed populations. The results of the test in Table B4 reject the null hypothesis for all of the comparative studies between SPM Decreasing Mutation and all the DPM which leads us to conclude that both sets of models solve the same problem by applying different criteria.

## 4.3.5 | Discussions

In Tables C5, C6 and C7, as well as in Figure 24, a comparative summary of the different scenarios studied is presented, considering for each of them the minimum, mean, and maximum values of the series of 31 executions.

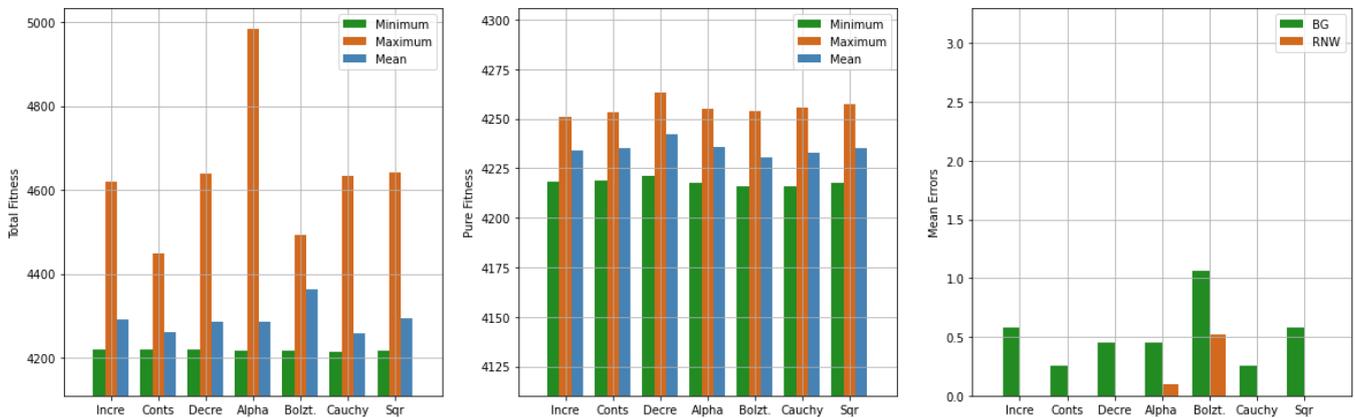

**FIGURE 24** Summary of the results obtained for every scenario studied

From these results, the following conclusions arise:

- The best results for Pure and Total Fitness are obtained using DPM with Cauchy criteria and a decreasing mutation factor.
- The model that presents the best behavior against both BG and RNW errors is SPM with constant mutation.
- The model that performs the worst is the SPM with decreasing mutation.
- The differences between the models are not very significant; for example, the differences in Pure Fitness between the best and the worst models are less than 0.13%.
- The normality study together with the homogeneity study ensure that, in general, all the different SPM models solve the same problem as do all the DPM models. However, the homogeneity studies show a difference between the models solved by SPM and DPM, this must be due to the critera applied in the weighting of the CE errors when obtaining the Total Fitness.

In order to select the best criteria for solving the model, the different scenarios are compared using a multi-criteria decision-making (MCDM): the Electre method (Figueira, Mousseau, & Roy, 2016). This method is considered an overclassification method that is based on the overqualification relationships that may exist between each two alternatives. An alternative 'a' is said to overqualify another alternative 'b', if 'a' is as good as 'b' in most of the criteria and not significantly worse in the rest of them.

The alternatives whose behaviors have been contrasted are the three series corresponding to different levels of mutation together with the four series of the DPM methods.

Major problems involve the selection of the attributes of each series to be taken into account, as well as the importance of each of them. We have taken into account a total of nine attributes divided into three categories: the fitness function (max, median, min, standard deviation) shown in Table D8; the BG assignment errors (max, median, standard deviation) shown in Table D9; and the runtime for each scenario (median and standard deviation).

Once the attribute selection problem has been solved, it remains to determine the weights of each one of them. The values corresponding to the median are those that have been considered the most important with the fitness function ahead of the BG errors and the runtimes (see row rating in Tables D8 and D9).

The result of the application of the Electre method, shown in Table D10, has yielded as the dominant data series the one corresponding to DPM Cauchy that beats the other six alternatives, not being overcome by any of them. The next two best series are SPM Constant Mutation and DPM Square root. On the other hand, the ones that show the worst behavior are the data series of SPM Decreasing mutation.

## 5 | CONCLUSIONS AND FUTURE WORK

In this work, we proposed a GA model to solve the Airport LTO optimization problem for the scheduling of both boarding gates and runways for arrivals and departures. Different mutation criteria and CHTs have been studied to find the best combination. The model was tested with a real-world scenario at the JFK airport for one day of flights, showing very promising results.

One of the successes of the model was the formulation of the gene structure where, by means of an integer number, all the necessary information has been included without hindering the programming or slowing down the calculations. From the point of view of the dimensions of the problem and thanks to the genetic coding used, no limitations are foreseen either to the number of ATMs or to the configuration of the airport.

Within the wide range of hyperparameters of the model, those that showed better behavior were one-point crossover, use of parent-offspring replacement through competition , constant mutation, and DPM for the constraint handling. From the evolutionary process analysis, it can be deduced that the optimization process, if the components of the fitness function are well weighted, always seeks to satisfy the boundary conditions during the first evolutionary levels to subsequently proceed with a continuous adjustment of the pure fitness reaching at the end of the process a value of Pure Fitness very close to the optimum obtained by binary linear programming.

The next steps in the development of this GA model include:

- Allowing the update of the optimal solution found for the daily ATMs based on delays/advances of some operations, providing dynamism to the process by allowing flexibility of assignments depending on daily operations real schedule.
- Introduction of the relationship between flight times and the times passengers take in large airports to reach the exit as an additional optimization factor.
- Including fuzzy logic criteria as part of the problem, either to guide the behavior of the hyperparameters of the problem or, as the case may be, to control the congestion problem that could be generated in both taxiways and runway headers (L. Li, Lin, & Liu., 2006; Koukol, Zajickova, Marek, & pavel Tucek., 2015).
- In addition, and outside the airport sector, a model such as this one could be perfectly extrapolated to sectors such as logistics where it would suffice to change Terminals for Logistics Centers, RNW for Nearby distribution centers and distances to RNW by distances between Logistics Centers and Proximity Centers.

## FUNDING SUPPORT

This work was supported by the Spanish Ministry of Science and Education within the framework of the FightDIS project (PID2020-H7263GB-100, 2021-2023); and the European Commission with the IBERIFIER project (2021-2023): Iberian Digital Media Research and FactChecking Hub (CEF-TC-2020-2: 2020-EU-IA-0252).

## CONFLICT OF INTEREST

The authors declare no potential conflict of interests.

# APPENDIX

## A  MUTATION AND PENALTY FACTOR SERIES GRAPHS

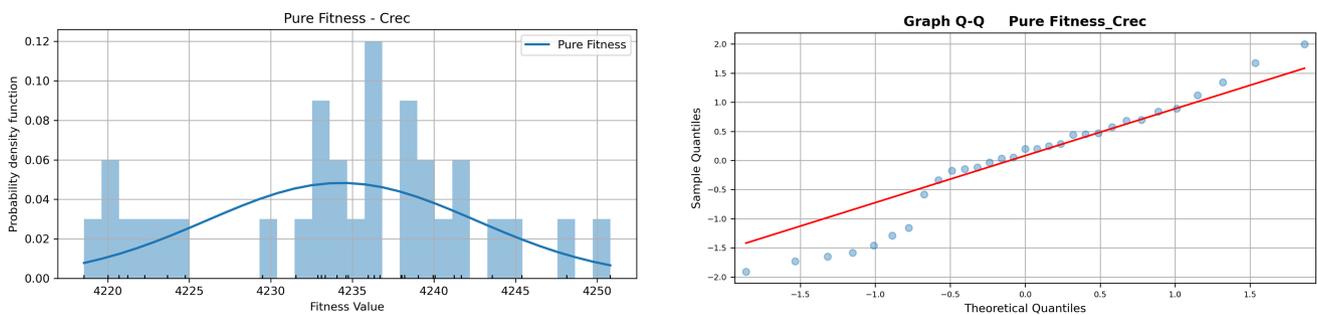

**FIGURE A1**  Increasing mutation factor

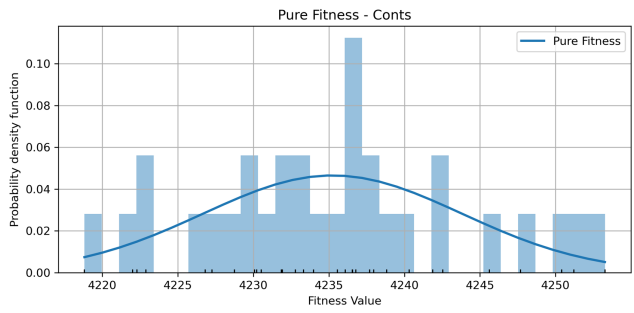 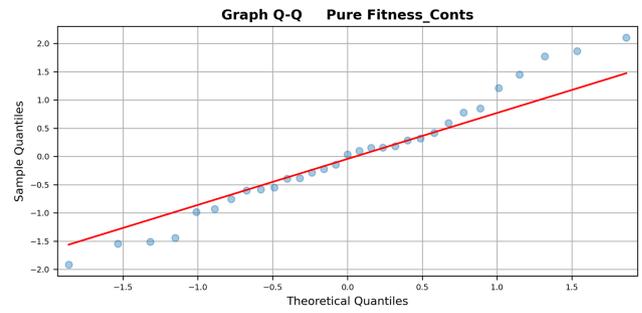

**FIGURE A2** Constant mutation factor

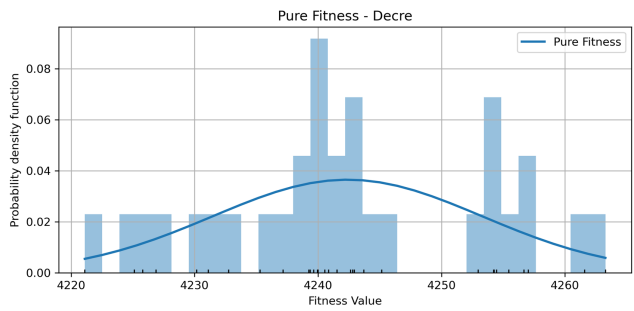 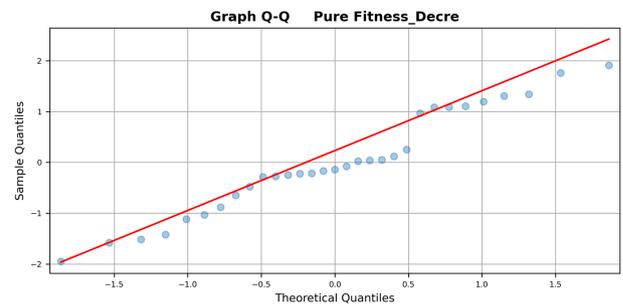

**FIGURE A3** Decreasing mutation factor

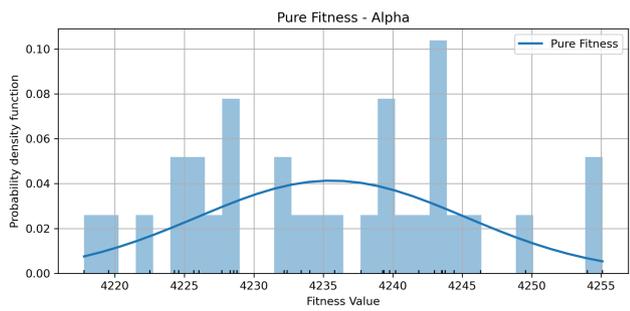 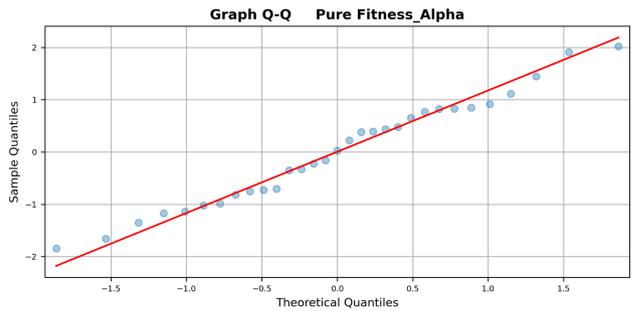

**FIGURE A4** Alpha penalty factor

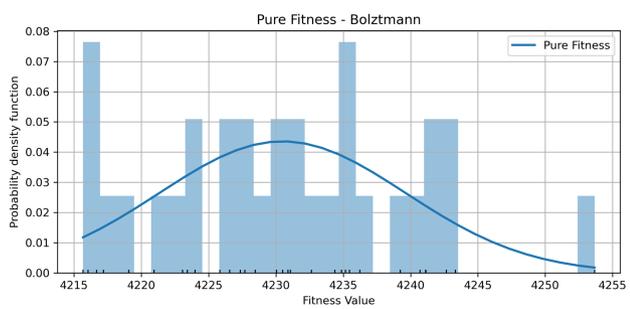 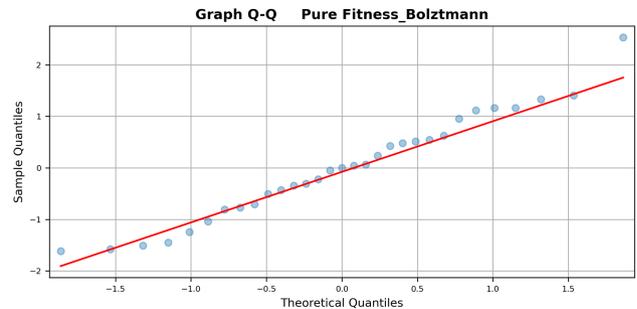

**FIGURE A5** Boltzmann penalty factor

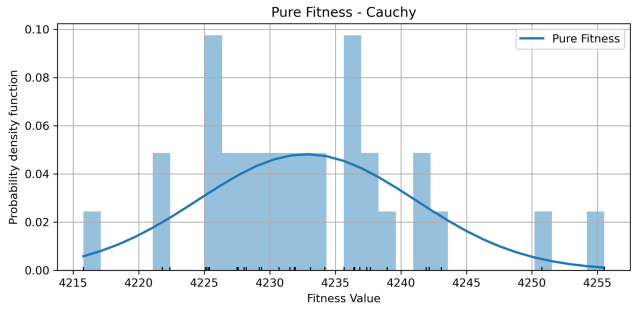 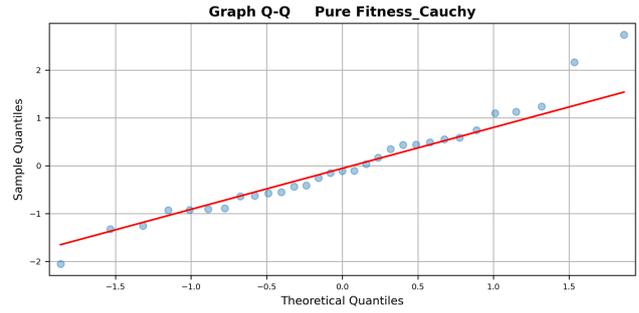

**FIGURE A6**   Cauchy penalty factor

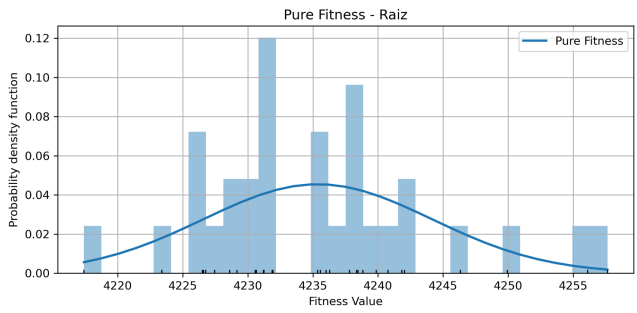 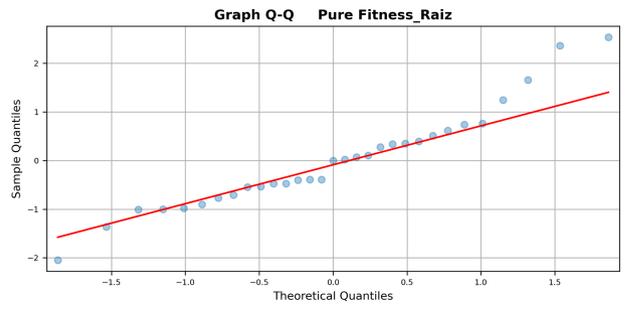

**FIGURE A7**   Square root penalty factor

# B NORMALITY AND P-VALUE TABLES

TABLE B1 ure Fitness Normality - Kurtosis and Skewness evaluation

| Fitness | Model | Kurtosis | Skewness | Result |
|---|---|---|---|---|
| Mutation | Increasing | -0.5597 | -0.2838 | Normal |
| | Constant | -0.4153 | 0.2412 | Normal |
| | Decreasing | -0.7483 | 0.0751 | Normal |
| Penalty Factors | Alpha | -0.8044 | 0.0931 | Normal |
| | Bolztmann | -0.3598 | 0.2618 | Normal |
| | Cauchy | 0.5826 | 0.6247 | Normal |
| | Square root | 0.5136 | 0.6778 | Normal |

TABLE B2 Pure Fitness Normality according to pvalue Shapiro / DAgostino

| Fitness | Model | Shapiro | DAgostino | H0 | Result |
|---|---|---|---|---|---|
| Mutation | Increasing | 0.2092 | 0.6870 | Accepted | Normal |
| | Constant | 0.7391 | 0.8074 | Accepted | Normal |
| | Decreasing | 0.3025 | 0.6675 | Accepted | Normal |
| Penalty Factors | Alpha | 0.6558 | 0.5752 | Accepted | Normal |
| | Bolztmann | 0.5998 | 0.7880 | Accepted | Normal |
| | Cauchy | 0.4357 | 0.1421 | Accepted | Normal |
| | Square root | 0.1924 | 0.1261 | Accepted | Normal |

TABLE B3 T-test over sets of Pure Fitness models

| Fitness | Model | Statistical | p-value | H0 | Result |
|---|---|---|---|---|---|
| Mutation | Incre - Const | -0.4222 | 0.6743 | Accepted | Same |
| | Incre - Decre | -3.2233 | 0.0021 | Rejected | Different |
| | Const - Decre | -2.8162 | 0.0066 | Rejected | Different |
| Penalty Factors | Decre - Alpha | 2.5538 | 0.0132 | Rejected | Different |
| | Decre - Boltzmann | 2.5538 | 0.0132 | Rejected | Different |
| | Decre - Cauchy | 3.8224 | 0.0003 | Rejected | Different |
| | Decre - Sqrt root | 2.7385 | 0.0081 | Rejected | Different |
| | Alpha - Boltzmann | 2.0967 | 0.0402 | Rejected | Different |
| | Alpha - Cauchy | 1.1935 | 0.2373 | Accepted | Same |
| | Alpha - Sqrt root | 0.0882 | 0.9300 | Accepted | Same |
| | Boltzmann - Cauchy | -1.0291 | 0.3075 | Accepted | Same |
| | Boltzmann - Sqr root | -2.1084 | 0.0392 | Rejected | Different |
| | Cauchy - Sqr root | -1.1624 | 0.2497 | Accepted | Same |

**TABLE B4** U-test over sets of Penalty factor models and decreasing mutation

| Fitness | Modelo | Homoc pval | H0 | UTest pval | H0 |
| --- | --- | --- | --- | --- | --- |
| Mutation | Incre - Const | 0.4470 | Accepted | 0.9438 | Accepted |
|  | Incre - Decre | 0.2721 | Accepted | 0.0029 | Rejected |
|  | Const - Decre | 0.6324 | Accepted | 0.0066 | Rejected |
| Penalty Factors | Decre - Alpha | 0.5927 | Accepted | 0.0291 | Rejected |
|  | Decre - Boltzmann | 0.1303 | Accepted | 0.0000 | Rejected |
|  | Decre - Cauchy | 0.7436 | Accepted | 0.0005 | Rejected |
|  | Decre - Sqrt root | 0.9168 | Accepted | 0.0078 | Rejected |
|  | Alpha - Boltzmann | 0.3268 | Accepted | 0.0412 | Rejected |
|  | Alpha - Cauchy | 0.3713 | Accepted | 0.1904 | Accepted |
|  | Alpha - Sqr root | 0.4808 | Accepted | 0.8000 | Accepted |
|  | Boltzmann - Cauchy | 0.0547 | Accepted | 0.3455 | Accepted |
|  | Boltzmann - Sqrt root | 0.0758 | Accepted | 0.0471 | Rejected |
|  | Cauchy - Sqr root | 0.7743 | Accepted | 0.2721 | Accepted |

## C  SERIAL VALUES: FITNESS AND ERRORS

**TABLE C5**  Minimum values sorted by Total Fitness

| Model | Total Fitness | Pure Fitness | BG Err | RNW Err |
|---|---|---|---|---|
| Cauchy | 4,215.79 | 4,215.79 | 0 | 0 |
| Bolztmann | 4,217.20 | 4,217.20 | 0 | 0 |
| Square root | 4,217.40 | 4,217.40 | 0 | 0 |
| Alpha | 4,217.78 | 4,217.78 | 0 | 0 |
| SPM Constant | 4,218.81 | 4,218.81 | 0 | 0 |
| SPM Increasing | 4,220.03 | 4,220.03 | 0 | 0 |
| SPM Decreasing | 4,221.11 | 4,221.11 | 0 | 0 |
| MEAN value | 4,218.11 | 4,218.11 | 0 | 0 |

**TABLE C6**  Mean values sorted by Total Fitness

| Model | Total Fitness | Pure Fitness | BG Err | RNW Err |
|---|---|---|---|---|
| Cauchy | 4,231.86 | 4,231.86 | 0 | 0 |
| Square root | 4,235.58 | 4,235.58 | 0 | 0 |
| SPM Constant | 4,236.55 | 4,236.55 | 0 | 0 |
| SPM Increasing | 4,237.98 | 4,238.98 | 0 | 0 |
| Alpha | 4,239.77 | 4,239.77 | 0 | 0 |
| SPM Decreasing | 4,242.63 | 4,242.63 | 0 | 0 |
| Bolztmenn | 4,331.06 | 4,231.06 | 1 | 0 |
| MEAN value | 4,250.78 | 4,236.49 | 0 | 0 |

**TABLE C7**  Maximum values sorted by Total Fitness

| Model | Total Fitness | Pure Fitness | BG Err | RNW Err |
|---|---|---|---|---|
| SPM Constant | 4,447.67 | 4,247.67 | 2 | 0 |
| Bolztmann | 4,493.34 | 4,243.34 | 2 | 1 |
| SPM Increasing | 4,618.55 | 4,218.55 | 4 | 0 |
| Cauchy | 4,634.21 | 4,234.21 | 4 | 0 |
| SPM Decreasing | 4,639.66 | 4,239.66 | 4 | 0 |
| Square root | 4,640.77 | 4,240.77 | 4 | 0 |
| Alpha | 4,983.41 | 4,233.41 | 6 | 0 |
| MEAN value | 4,636.80 | 4,236.80 | 3.7 | 0.6 |

# D  ELECTRE METHOD

**TABLE D8**  Decision matrix - Fitness function attributes

| ATTRIBUTES | Min | Median | Max | Std |
|---|---|---|---|---|
| RATING | 8 | 9 | 4 | 3 |
| Criteria (Min = 1; Max = 0) | 1 | 1 | 1 | 1 |
| Weighting factor | 17.8% | 20.0% | 8.9% | 6.7% |
| SPM Mutation Increasing | 4,220.03 | 4,237.98 | 4,218.55 | 8.254 |
| SPM Mutation Constant | 4,218.81 | 4,236.55 | 4,247.67 | 8.574 |
| SPM Mutation Decreasing | 4,221.11 | 4,242.61 | 4,239.66 | 10.926 |
| DPM Alpha | 4,217.78 | 4,239.77 | 4,233.41 | 9.654 |
| DPM Boltzmann | 4,217.20 | 4,231.06 | 4,243.34 | 9.157 |
| DPM Cauchy | 4,215.79 | 4,231.86 | 4,234.21 | 8.289 |
| DPM Square root | 4,217.40 | 4,235.58 | 4,240.77 | 8.781 |

**TABLE D9**  Decision matrix - BG errors and execution time attributes

| ATTRIBUTES | BG_Max | BG_Med | BG_Std | Time_Med | Time_Std |
|---|---|---|---|---|---|
| RATING | 6 | 7 | 5 | 2 | 1 |
| Criteria | 1 | 1 | 1 | 1 | 1 |
| Weighting factor | 13.3% | 15.6% | 11.1% | 4.4% | 2.2% |
| SPM Mutation increasing | 4 | 0 | 1.040 | 9.050 | 0.170 |
| SPM Mutation constant | 2 | 0 | 0.670 | 9.070 | 0.260 |
| SPM Mutation decreasing | 4 | 0 | 0.980 | 9.270 | 0.760 |
| DPM Alpha | 6 | 0 | 1.210 | 9.270 | 0.260 |
| DPM Boltzmann | 2 | 1 | 0.880 | 11.020 | 0.830 |
| DPM Cauchy | 4 | 0 | 0.840 | 8.980 | 0.080 |
| DPM Square root | 4 | 0 | 1.040 | 9.010 | 0.130 |

**TABLE D10**  Results Electre method - Aggregate Dominance matrix

| ATTRIBUTES | $\triangle$ | = | $\triangledown$ | $\alpha$ | $\beta$ | $\zeta$ | $\sqrt{()}$ | Beats |
|---|---|---|---|---|---|---|---|---|
| SPM Mutation increasing | 0 | 0 | 1 | 0 | 0 | 0 | 0 | 1 |
| SPM Mutation constant | 0 | 0 | 1 | 1 | 0 | 0 | 0 | 2 |
| SPM Mutation decreasing | 0 | 0 | 0 | 0 | 0 | 0 | 0 | 0 |
| DPM Alpha | 0 | 0 | 1 | 0 | 0 | 0 | 0 | 1 |
| DPM Boltzmann | 0 | 0 | 0 | 0 | 0 | 0 | 0 | 0 |
| DPM Cauchy | 1 | 1 | 1 | 1 | 1 | 0 | 1 | 6 |
| DPM Square root | 0 | 0 | 1 | 1 | 0 | 0 | 0 | 2 |
| Is overcomed | 1 | 1 | 5 | 3 | 1 | 0 | 1 | |